\documentclass[10pt,twocolumn,letterpaper]{article}

\usepackage{cvpr}              

\usepackage[dvipsnames]{xcolor}

\usepackage[accsupp]{axessibility} 
\usepackage{subcaption}
\usepackage{graphicx}
\usepackage{threeparttable}
\usepackage{amsmath}
\usepackage{amssymb}
\usepackage{booktabs}
\usepackage{multirow}
\usepackage{algorithm}
\usepackage{algpseudocode}
\usepackage{color, colortbl}
\usepackage{subcaption}
\definecolor{cvprblue}{rgb}{0.21,0.49,0.74}

\definecolor{Gray}{gray}{0.9}
\definecolor{Red}{RGB}{230, 57, 70}
\usepackage[pagebackref,breaklinks,colorlinks,citecolor=cvprblue]{hyperref}
\newcommand{\myPara}[1]{\vspace{.05in}\noindent\textbf{#1}}
\def\paperID{3966} 
\def\confName{CVPR}
\def\confYear{2024}

\title{Open-World Human-Object Interaction Detection via Multi-modal Prompts }

\author{Jie Yang$^{1,2}$\thanks{Equal contribution.}~~,~Bingliang Li$^{1}$$\footnote[1]{}$~~,~Ailing Zeng$^{2}$\thanks{Corresponding author.}~~, Lei Zhang$^{2}$, Ruimao Zhang$^{1}$\footnote[2]{}~~ \\
$^{1}$The Chinese University of Hong Kong, Shenzhen~~$^{2}$International Digital Economy Academy 
}

\begin{document}
\maketitle
\begin{abstract}

In this paper, we develop \textbf{MP-HOI}, a powerful \textbf{M}ulti-modal \textbf{P}rompt-based \textbf{HOI} detector designed to
leverage both textual descriptions for open-set generalization and visual exemplars for handling high ambiguity in descriptions, realizing HOI detection in the open world. 
Specifically, it integrates visual prompts into existing language-guided-only HOI detectors to handle situations where textual descriptions face difficulties in generalization and to address complex scenarios with high interaction ambiguity.
To facilitate MP-HOI training, we build a large-scale HOI dataset named Magic-HOI, which gathers six existing datasets into a unified label space, forming over 186K images with 2.4K objects, 1.2K actions, and 20K HOI interactions. Furthermore, to tackle the long-tail issue within the Magic-HOI dataset, we introduce an automated pipeline for generating realistically annotated HOI images and present SynHOI, a high-quality synthetic HOI dataset containing 100K images. 
Leveraging these two datasets, MP-HOI optimizes the HOI task as a similarity learning process between multi-modal prompts and objects/interactions via a unified contrastive loss, to learn generalizable and transferable objects/interactions representations from large-scale data.
MP-HOI could serve as a generalist HOI detector, surpassing the HOI vocabulary of existing expert models by more than 30 times. 
Concurrently, our results demonstrate that MP-HOI exhibits remarkable zero-shot capability in real-world scenarios and consistently achieves a new state-of-the-art performance across various benchmarks. Our project homepage is available at \url{https://MP-HOI.github.io/}.

\end{abstract}    
\section{Introduction}
\label{sec:Introduction}
Human-Object Interaction Detection (HOI), a core element in human-centric vision perception tasks such as human activity recognition~\cite{heilbron2015activitynet}, motion tracking~\cite{yi2022physical}, and anomaly behavior detection~\cite{liu2018future}, has attracted considerable attention over the past decades. 
HOI primarily involves localizing correlated human-object 2D positions within an image and identifying their interactions.
Although numerous models have been proposed~\cite{hou2020visual,wan2019pose,liao2022gen,wu2022mining,zhang2021mining,zhou2022human}, 
deploying these models in open-world scenarios remains a significant challenge due to their primary training on closed-set data and limited generalization capabilities.

\begin{figure*}[t]
    \centering
    \includegraphics[width=0.9\linewidth]{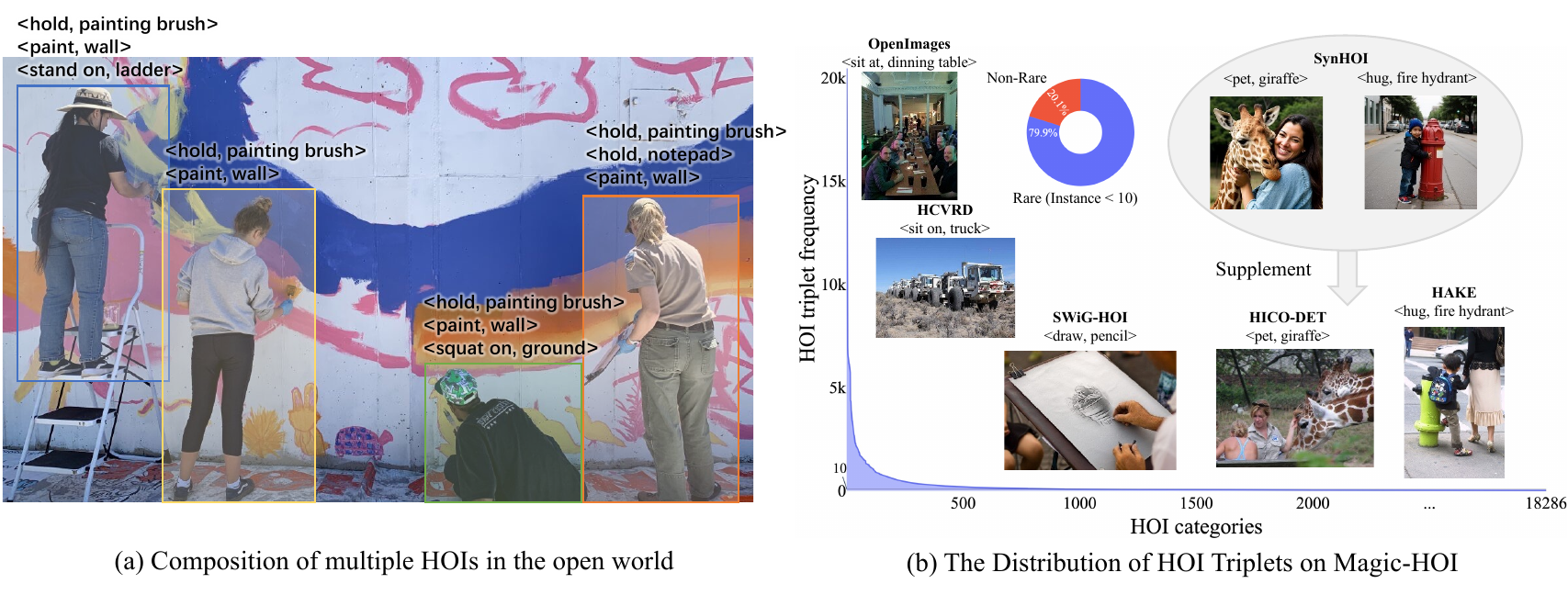}
    \vspace{-0.5cm}
    \caption{We show (a) the coexisting composited interactions within the same person in an in-the-wild (e.g., A man is squatting on the ground, holding a painting brush, and painting the wall); (b) the long-tail distribution issue in our \emph{Magic-HOI} dataset, along with another proposed \emph{SynHOI} dataset to address it.}
            \vspace{-0.1cm}
    \label{fig:intro}
    \vspace{-0.5cm}
\end{figure*}

With the advancements in visual-linguistic models, such as CLIP~\cite{radford2021learning}, recent research~\cite{liao2022gen,wu2022end,ning2023hoiclip} attempts to introduce the natural language into HOI task and transfer knowledge from CLIP to recognize unseen HOI concepts. However, these methods still suffer from the following limitations:
(1) \textbf{Limited data and category definitions}: They primarily rely on training data from a single dataset, such as HICO-DET~\cite{chao:wacv2018}, which is built upon a relatively small set of predefined objects (e.g., $80$) and actions (e.g., $117$). 
Consequently, this narrow range of object and action categories restricts their ability to generalize effectively. 
(2) \textbf{An inherent bottleneck with textual descriptions}: Even when the model is expanded to be trained on more objects and action categories, it still encounters a bottleneck when dealing with entirely new categories that lack any relevant text descriptions in the training data.
(3) \textbf{Difficulty addressing high-ambiguity scenarios}: Current methods typically associate a human with a specific object based solely on a single interaction description. However, real-world situations often involve the composition of multiple interactions, as depicted in Fig~\ref{fig:intro}-(a), introducing significant ambiguity in interaction learning. Moreover, these models may fail to detect a person engaged in a specific combination of multiple HOIs, which is essential for certain practical applications.

To tackle the above issues, we develop \emph{MP-HOI}, a novel and powerful Multi-modal Prompt-based HOI generalist model for effective HOI detection in the open world. Within \emph{MP-HOI},
we incorporate the visual prompts into existing language-guided-only HOI detectors to enhance the effectiveness and generalization capability. 
Visual prompts can act as exemplars, allowing the model to detect the same objects/interactions (single HOI or the combination of multiple HOIs for the same person), and they also aid in reducing semantic ambiguity present in textual prompts. Simultaneously, textual prompts also provide semantic context to enhance the understanding of visual prompts.
The proposed \emph{MP-HOI} optimize the
HOI task as a similarity learning process between multi-modal prompts and objects/interactions by using a unified cross-modality contrastive loss. It
allows the model to learn the textual/visual prompts-to-objects/interactions alignments from large-scale data.

To facilitate the effective training of our model, we build a large-scale HOI dataset named \emph{Magic-HOI}, which gathers six existing datasets into a unified label space, forming the one containing over $186K$ images with $2.4K$ objects, $1.2K$ actions, and $20K$ HOI interactions. 
Furthermore, as shown in Fig.~\ref{fig:intro}-(b), we analyze the distribution of all HOI interactions in \emph{Magic-HOI}, finding a significant long-tail distribution. Upon further investigation, we discover that $79.9\%$ of HOI interactions have fewer than $10$ instances, and over $5900$ HOI categories have only a single image available. Such imbalanced label distribution challenges prompt-based learning, particularly visual prompts. 
Unfortunately, our further investigation reveals that rare HOI categories are genuinely rare in real-life scenarios, making it difficult to collect data from the internet.

Inspired by the impressive conditioned image generation capabilities demonstrated by recent text-to-image diffusion models~\cite{balaji2022ediffi,zhou2022towards,saharia2022photorealistic,ramesh2022hierarchical,rombach2022high}, 
we present a diffusion-based automatic pipeline, including the HOIPrompt design, automatic labeling and filtering, and quality verification, designed to systematically scale up the generation of diverse and high-precision HOI-annotated data. Based on it, we present \emph{SynHOI}, a high-diversity synthetic HOI dataset with over $100K$ fully annotated HOI images, which can fill in the shortcomings of the long-tail \emph{MP-HOI} dataset.

By leveraging the newly collected \emph{MP-HOI} and \emph{SynHOI} datasets, we train the \emph{MP-HOI} to be a generalist HOI detector, surpassing the HOI vocabulary of existing expert models (e.g., GEN-VLKT~\cite{liao2022gen}) by more than $30$ times. Through extensive experiments, we demonstrate that \emph{MP-HOI} exhibits remarkable zero-shot capability in real-world scenarios. Furthermore, \emph{MP-HOI} consistently achieves the new state-of-the-art performance across various benchmarks.
The main contributions of this paper are threefold.

(1) We develop the first multi-modal prompts-based generalist HOI detector, called \emph{MP-HOI}, which leverages
both textual descriptions for open-set generalization and visual exemplars for handling high ambiguity in descriptions, paving a new path to HOI detection in the open world.

(2) We build a large-scale \emph{Magic-HOI} dataset for effective large-scale training, which gathers six existing datasets into a unified semantic space.
To address the long-tail issue in \emph{Magic-HOI}, we construct an automated pipeline for generating
realistically annotated HOI images and present \emph{SynHOI},
a high-quality synthetic HOI dataset.

(3) Extensive experimental results demonstrate that the proposed \emph{MP-HOI} exhibits remarkable zero-shot capability in real-world scenarios and consistently achieves a new state-of-the-art performance across various benchmarks.

\section{Related Work}
\textbf{HOI Detection.}
HOI detection task primarily encompasses three sub-problems, \textit{i.e.}, object detection, human-object pairing, and interaction recognition.
Previous HOI detectors can generally be divided into one-stage and two-stage paradigms.
The two-stage strategy employs an off-the-shelf detector to determine the locations and classes of objects, followed by specially-designed modules for human-object association and interaction recognition.
Most methods are dedicated to exploring additional feature streams to improve interaction classification, such as the appearance stream ~\cite{gao2018ican,li2019transferable,kim2020uniondet,hou2021detecting}, spatial stream~\cite{xu2019learning,bansal2020detecting,li2020detailed}, pose and body-parts stream~\cite{gupta2019no,wan2019pose,li2020detailed}, semantic stream~\cite{liu2020consnet,bansal2020detecting,gao2020drg}, and graph network~\cite{qi2018learning,xu2019learning,wang2020contextual,zhang2021spatially,park2023viplo}.
Instead, the one-stage strategy detects HOI triplets in a single forward pass by assigning human and object proposals to predefined anchors and then estimating potential interactions~\cite{liao2020ppdm, wang2020learning,kim2020uniondet,fang2021dirv}. 
Recently, the DETR-based HOI detectors~\cite{Tamura_2021_CVPR,tamura2021qpic,kim2021hotr,chen2021qahoi,ma2023fgahoi} have gained prominence in this paradigm, which
formulate the HOI detection task as a set prediction problem, avoiding complex post-processing. In particular, many methods~\cite{zhang2021mining,liao2022gen,zhou2022human,chen2021reformulating,ning2023hoiclip,cao2023re,yuan2022rlip,yuan2023rlipv2,kim2023relational,zhao2023unified} demonstrate promising performance improvements by disentangling
human-object detection and interaction classification as two decoders in a cascade manner. Our work builds on the transformer-based HOI detection strategy and focuses on developing the first multi-modal prompts-based generalist HOI detector.

\myPara{Zero-shot HOI Detection.}
Zero-shot HOI detection has emerged as a field aiming to identify unseen HOI triplet categories not present in the training data.
Previous research~\cite{bansal2020detecting,peyre2019detecting,peyre2019detecting,gupta2019no,hou2020visual,hou2021affordance,hou2021detecting,liu2020consnet} has addressed this task in a compositional manner, by disentangling the reasoning process on actions and objects during training. This approach enables the recognition of unseen HOI triplets during inference.
With the advancements in Vision-Language Models, such as CLIP~\cite{radford2021learning}, recent research~\cite{liao2022gen,wu2022end,ning2023hoiclip} has shifted focus toward transferring knowledge from CLIP to recognize unseen HOI concepts. This shift has resulted in a notable performance improvement in zero-shot settings. Our work aims to develop a HOI detector further to generalize in open-world scenes effectively.

\section{Datasets}
\label{sec:data}
This section introduces the proposed two datasets: 1) \emph{Magic-HOI}: A unified large-scale HOI dataset that gathers six existing datasets to form a unified one with consistent label space; 2) \emph{SynHOI}: A high-quality synthetic HOI dataset to tackle the long-tail issue within \emph{Magic-HOI}.

\begin{table}[t]
\begin{center}
\resizebox{\linewidth}{!}{
		\begin{tabular}{l|ccc|cc}
			\hline
			Datasets & Objects & Interactions & HOIs & Unified Images &  Unified Instances \\ \hline
            HICO-DET~\cite{chao:wacv2018} & 80 & 117  & 600  & 37,633  & 117,871 \\
            SWiG-HOI~\cite{wang2021discovering} & 1,000 & 407  &  13,520 & 50,861  &  73,499 \\
            OpenImages~\cite{kuznetsova2020open} & 80 & 117  & 600 & 20,140 & 78,792 \\
            PIC~\cite{PicDataset_2023} & 80 & 117  & 600 & 2,736 & 9,503 \\
            HCVRD~\cite{zhuang2018hcvrd} & 1,821 & 884  & 6,311 & 41,586 & 256,550 \\
            HAKE~\cite{li2019hake} & 80 & 117  & 600 & 33,059 & 91,120 \\  
            \hline
           \emph{Magic-HOI}  & 2,427 & 1,227 & 20,036 & 186,015 & 627,335 \\ \hline
		\end{tabular}   
}
\end{center}
\vspace{-0.5cm}
    \caption{Statistics of \emph{Magic-HOI} that unifies six existing datasets. Displayed numerically within the table are the counts of attributes we have integrated from each source dataset.} 
\label{tab:magichoi}
\vspace{-0.5cm}
\end{table}
\subsection{Magic-HOI}

Existing HOI detection datasets have incorporated a wide range of object and interaction categories. However, current models are typically trained on separate datasets, resulting in limited generalization capabilities.
In light of this, as shown in Tab.~\ref{tab:magichoi}, we propose \emph{Magic-HOI} by unifying six HOI detection datasets, including HICO-DET~\cite{chao:wacv2018}, SWiG-HOI~\cite{wang2021discovering}, OpenImages~\cite{kuznetsova2020open}, PIC~\cite{PicDataset_2023}, HCVRD~\cite{zhuang2018hcvrd}, HAKE~\cite{li2019hake}. To ensure the integrity of \emph{Magic-HOI}, we meticulously curate the object and verb labels across these datasets, diligently addressing issues of label overlap and potential data leakage for zero-shot testing. \emph{Magic-HOI} offers three key advantages: \textbf{(1) Rich synonymous descriptions:} Traditional datasets often use specific descriptions to annotate objects and relationships. In contrast, \emph{Magic-HOI} provides a multitude of synonymous descriptions across datasets. For example, it distinguishes between the object `person' and `woman'/`man', as well as the action descriptions `read' and `look at'. This diversity enhances the model's robustness and adaptability. 
\textbf{(2) Large-scale data:} Training large models necessitates diverse data. Single datasets, such as the $20K$ images in HICO-DET, can cause the model underfitting. \emph{Magic-HOI} make a significant effort to build a large-scale dataset for effective training, containing an extensive over $186K$ images.
\textbf{(3) Rich object/actions/interactions categories:} Current models are primarily trained on data with $80$ objects and $117$ verbs. However, \emph{Magic-HOI} significantly broadens the scope by offering over $2.4K$ objects, $1.2K$ actions, and $20K$ HOI interactions. Such a large vocabulary greatly enhances the model's generalization capabilities.

\begin{figure}[t]
  \centering
  \begin{subfigure}[b]{0.45\textwidth}
  \centering
    \includegraphics[width=\textwidth]{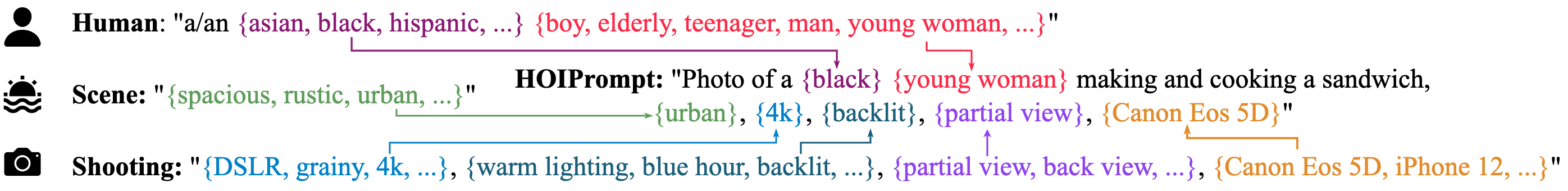}
    \caption{Examples of HOIPrompts}
    \label{fig:hoiprompt}
  \end{subfigure}
  \vfill
  \vspace{0.1cm}
    \centering
  \begin{subfigure}[b]{0.4\textwidth}
  \centering
    \includegraphics[width=\textwidth]{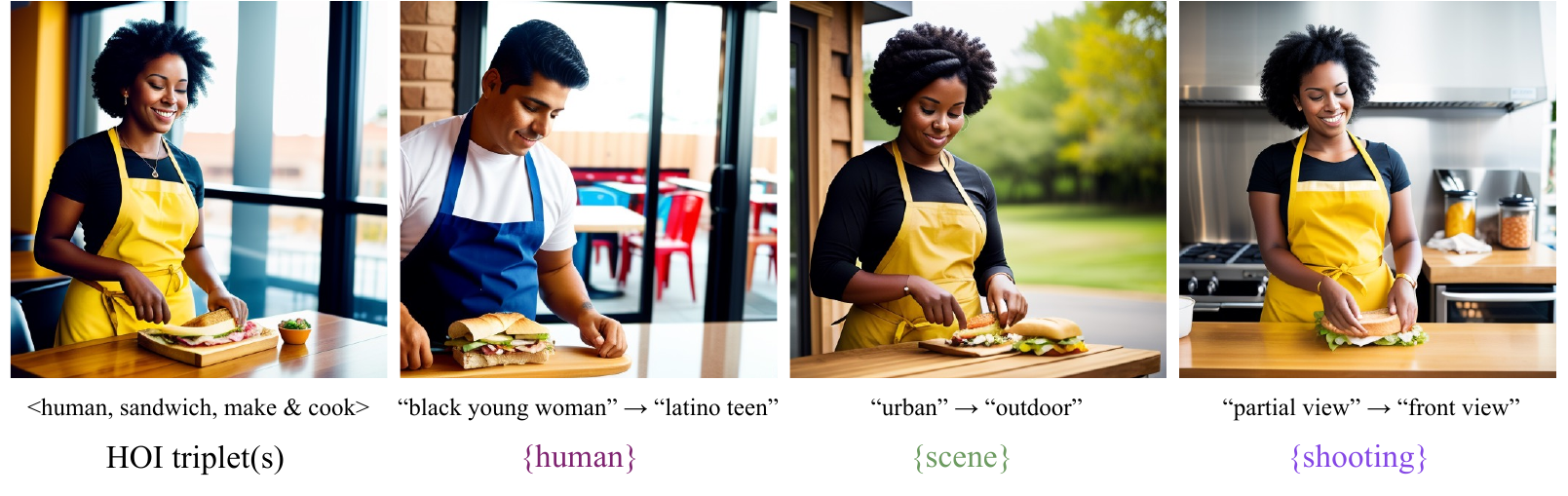}
    \caption{Examples of generated images and annotations}
    \label{fig:example_img}
  \end{subfigure}
    \vspace{-0.1cm}
  \caption{Illustration of a) HOIPrompts and b) how HOIPrompts guide the text-to-image generation process to enhance diversity. For more visualization, please refer to the Appendix. }
  \vspace{-0.6cm}
  \label{fig:data}
\end{figure}

\subsection{SynHOI}

As shown in Fig.~\ref{fig:intro}-(b), \emph{Magic-HOI} exhibits an inherent long-tail issue. Thus, we present a high-quality synthetic HOI dataset called \emph{SynHOI} to complement \emph{Magic-HOI}. To make the flow of dataset production scalable, we present an automatic pipeline, including the HOIPrompt design, automatic labeling and filtering, and quality verification, designed to continually scale up the generation of diverse and high-precision HOI-annotated data, as shown in Fig.~\ref{fig:data}. Please refer to the Appendix for details.
\begin{figure*}[t]
    \centering
    \includegraphics[width=\linewidth]{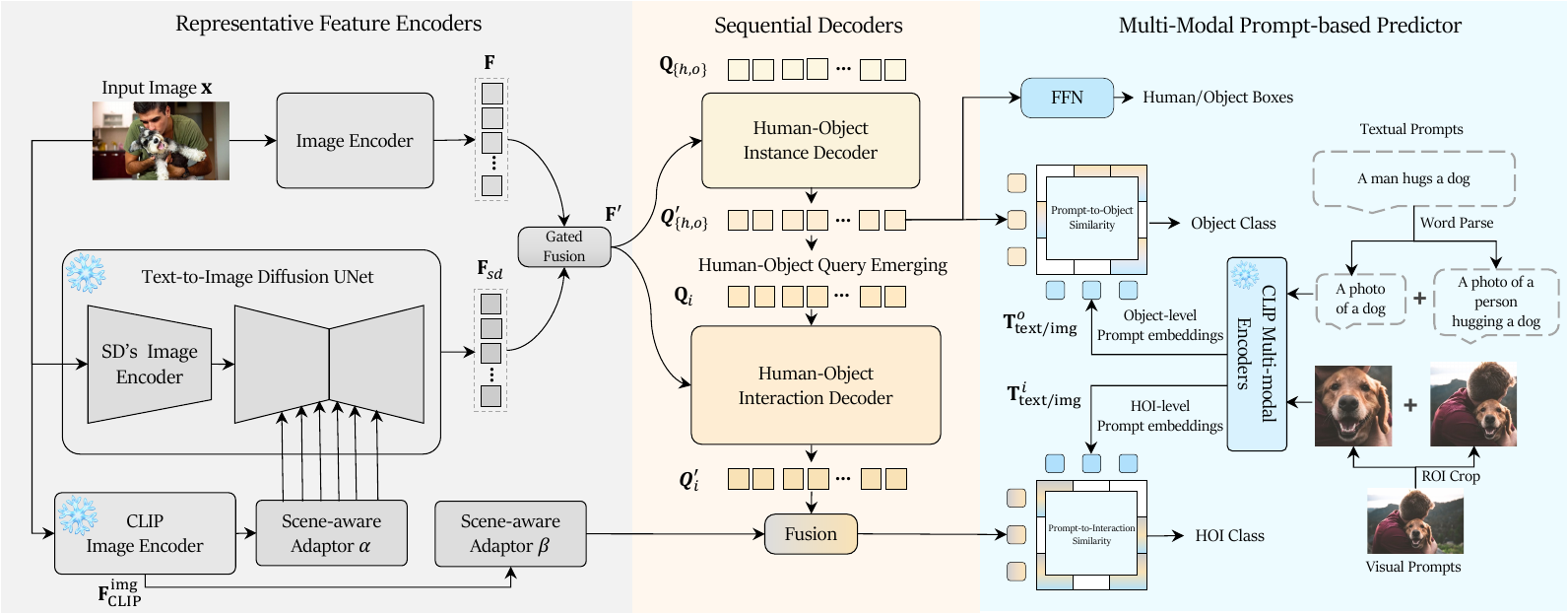}
    \caption{Overview of \emph{MP-HOI}, comprising three components: Representative Feature Encoder, Sequential Instance and Interaction Decoders, and Multi-modal Prompt-based Predictor. Ultimately, it can leverage textual or visual prompts to detect open-world HOIs.
    }
    \label{fig:framework}
    \vspace{-0.5cm}
\end{figure*}

\emph{SynHOI} has three key data characteristics:
(1) \textbf{High-quality data.} \emph{SynHOI} showcases high-quality HOI annotations. First, we employ CLIPScore~\cite{hessel2021clipscore} to measure the similarity between the synthetic images and the corresponding HOI triplet prompts. The \emph{SynHOI} dataset achieves a high CLIPScore of $0.849$, indicating a faithful reflection of the HOI triplet information in the synthetic images. Second, Fig.~\ref{fig:data}-(b) provides evidence of the high quality of detection annotations in \emph{SynHOI}, attributed to the effectiveness of the SOTA detector~\cite{zhang2022dino} and the alignment of \emph{SynHOI} with real-world data distributions. 
The visualization of \emph{SynHOI} is presented in the Appendix.
(2) \textbf{High-diversity data.} 
\emph{SynHOI} exhibits high diversity, offering a wide range of visually distinct images. Fig.~\ref{fig:data}-(b) demonstrates the impact of random variations in person's descriptions, environments, and photographic information within the HOIPrompts on the diversity of synthetic images.
(3) \textbf{Large-scale data with rich categories.} \emph{SynHOI} aligns \emph{Magic-HOI}'s category definitions to effectively address the long-tail issue in \emph{Magic-HOI}. It consists of over $100K$ images, $130K$ person bounding boxes, $140K$ object bounding boxes, and $240K$ HOI triplet instances.

\section{Methodology}
\subsection{The Overview of MP-HOI}
\label{sec:overview}
\emph{MP-HOI} is a multi-modal prompt-based HOI detector comprising three components: Representative Feature Encoder, Sequential Instance and Interaction Decoders, and Multi-modal Prompt-based Predictor, as shown in Fig.~\ref{fig:framework}. 

Given an input image $\textbf{x}$, the Representative Feature Encoder firstly encodes it to powerful feature representations $\textbf{F}^{\prime}$, which have fused the vanilla image features $\textbf{F}$ from the image encoder and the HOI-associated representations $\textbf{F}_{sd}$ obtained from text-to-image diffusion model~\cite{rombach2022high} (refer to Sec.~\ref{sec:StableDiffusion} for details). Based on $\textbf{F}^{\prime}$, the Sequential Instance and Interaction Decoders are designed to decode the pair-wise human-object representations and their corresponding interaction representations in a sequential manner. Specifically, the instance decoder takes the input of initialized pair-wise human-object queries $\boldsymbol{Q}_{\{h,o\}} \in \mathbb{R}^{2  \times N \times C }$ and produces the updated queries $\boldsymbol{Q}^{\prime}_{\{h,o\}} \in \mathbb{R}^{2  \times N \times C }$, where $N$ is the number of paired queries and $C$ indicates the channel dimension. Subsequently, we merge the human-object queries $\boldsymbol{Q}^{\prime}_{\{h,o\}}$, into interaction queries $\boldsymbol{Q}_{i} \in \mathbb{R}^{ N \times C }$ via average pooling. The interaction decoder then receives these interaction queries and updates them to $\boldsymbol{Q}^{\prime}_{i} \in \mathbb{R}^{ N \times C }$.
To enhance the interaction representations, we integrate the global CLIP image representation into
each of the interaction queries $\boldsymbol{Q}^{\prime}_{i} \in \mathbb{R}^{ N \times C }$ via element-wise addition.

The complete HOI prediction for an image includes human and object bounding boxes, object class, and interactions class. Firstly, we employ Feed-Foreward Network (FFN) on $\boldsymbol{Q}^{\prime}_{h,o}$ to predict the human and object bounding box. Secondly, we introduce the Multi-modal Prompt-based Predictor to identify the object and interaction representations into object and interaction classes (refer to Sec.~\ref{sec:CLIP}).

\subsection{Representative Feature Encoder}
\label{sec:StableDiffusion}

\begin{figure}[t]
    \centering
    \includegraphics[width=\linewidth]{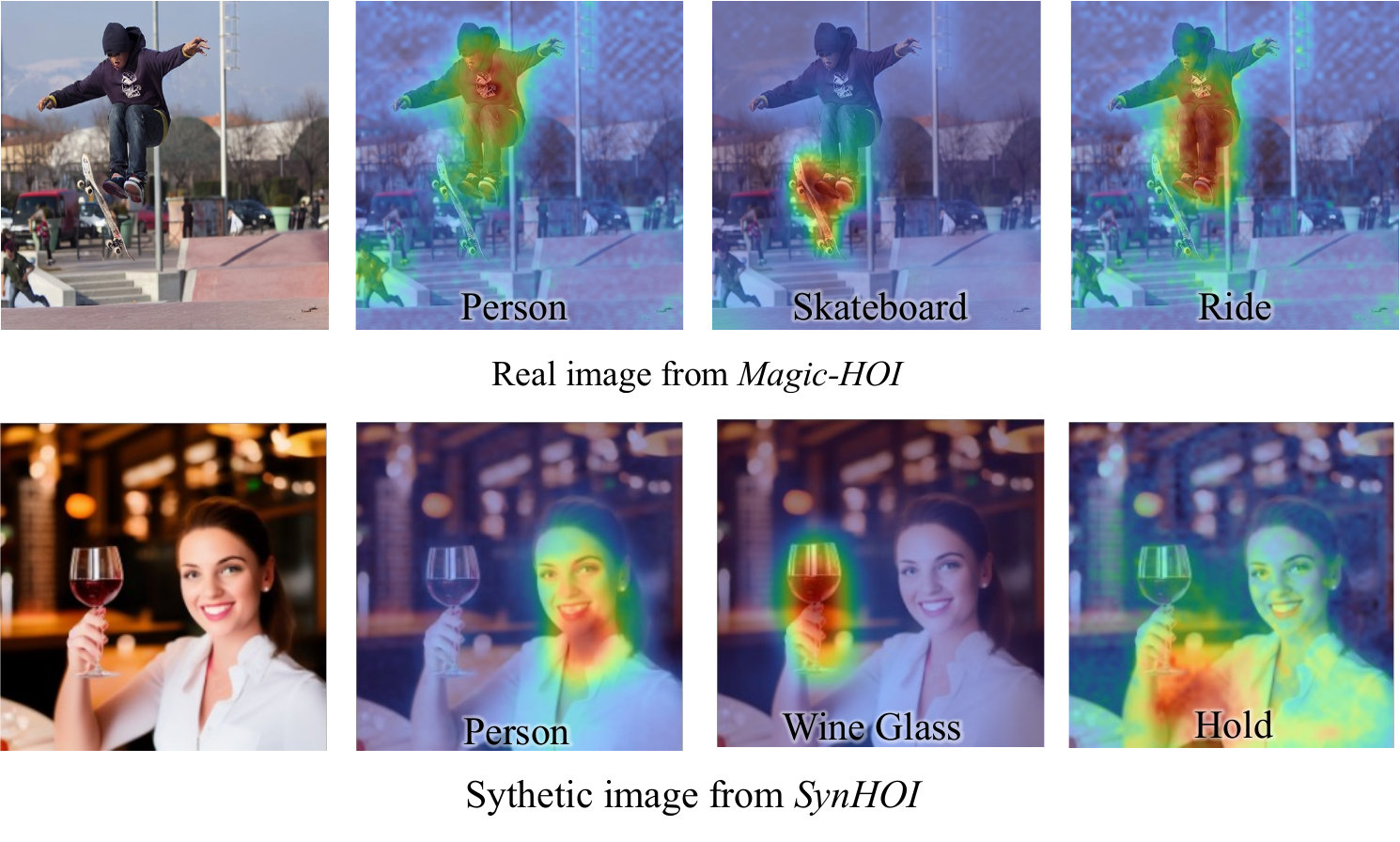}
    \vspace{-0.8cm}
    \caption{Attribution analysis of Stable Diffusion between object/interaction texts and real/synthetic images. The visualization solely utilizes time-step $0$.}
            \vspace{-0.5cm}
    \label{fig:daam}
    \vspace{-0.2cm}
\end{figure}

\textbf{Motivation.} As illustrated in Sec.~\ref{sec:data}, even though we have unified six datasets into \emph{Magic-HOI} and introduced the synthetic dataset \emph{SynHOI}, compared with the amount of training data for large-scale visual-linguistic models, such as CLIP~\cite{radford2021learning} or Stable Diffusion (SD)~\cite{rombach2022high}, we still have $1,000$ times less data. 
We firmly believe that such models can further provide effective encoded image representations for training instance/interaction decoders.
Specifically, considering that the SD model utilizes cross-attention mechanisms between text embeddings and visual representations for text-conditioned image generation, we assume the SD model has a substantial correlation between their feature spaces and semantic concepts in language. To further demonstrate this, we employ DAAM~\cite{tang2022daam} to generate pixel-level linguistic concept heatmaps based on the image features extracted from the SD model. As depicted in Fig.~\ref{fig:daam}, beyond the noun concepts highlighted in prior studies~\cite{xu2023open,li2023guiding,shiparddiversity},
we find that the internal representation space of a frozen SD model is highly
relevant to verb concepts and their associated contexts, which are very useful for the HOI task.  

\myPara{Extract Local Semantic Association in Stable Diffusion.}
Inspired by the above observations, the feature maps derived from the SD model are expected to exhibit superior semantic correspondence within local regions, establishing an association between the semantic information of the text and the corresponding regions within the image. 
Accordingly, we utilize the UNet-based diffusion model to extract the feature maps, which are not only associated with the noun concepts but also include the verb concepts and the corresponding contextual details.
Different from using a series of denoising steps to generate an image, we directly feed the input image $\boldsymbol{x}$ into the UNet and perform a single forward pass via the network. 
The output multi-scale image features are as $\boldsymbol{F}_{sd} = {\rm{SD}}(\boldsymbol{x},\boldsymbol{A}_x)$, 
where $\boldsymbol{A}_x$ denotes the text-related representation that corresponds to $\boldsymbol{x}$.
Typically, $\boldsymbol{A}_x$ can be obtained by utilizing a text encoder, such as the CLIP text encoder, to encode the description of the image. 
However, as a discriminative task, the description of the HOI image is not available in the inference.
To address this, we replace it with the global feature representation of $\boldsymbol{x}$ extracted from the CLIP image encoder, which bears significant relevance to the textual representation.

\myPara{Two Scene-aware Adaptors.} 
Given that the CLIP image features $\textbf{F}_{clip}^{img}$ represents a global feature, it inherently encompasses the contextual information of the overall scene.
As shown in Fig.~\ref{fig:framework}, we introduce two scene-aware adaptors, denoted as $\alpha$ and $\beta$, to project $\textbf{F}_{clip}^{img}$ into feature spaces more consistent with the SD model and Multi-modal Prompt-based Predictor.
Regarding the scene-aware adaptor $\alpha$, since the CLIP model is trained to align global visual and textual representations within a latent space, the $\textbf{F}_{clip}^{img}$ can be employed as a substitute for the textual representation. Hence, we can train an MLP to directly project $\textbf{F}_{clip}^{img}$ to a textual space suitable for the SD model.
As for the scene-aware adaptor $\beta$, we project $\textbf{F}_{clip}^{img}$ through it and incorporate the adapted version into each interaction query of $\boldsymbol{Q}^{\prime}_i$.
This adjustment allows for the tuning of these interaction queries to align more effectively with multi-modal prompts.

\myPara{Fused Features for Better HOI Representing.} Following the previous detectors~\cite{liao2022gen,zhang2021mining}, we also employ a common image encoder to obtain the vanilla image features $\textbf{F}$. Then, we perform Gated Fusion to generate more informative $\textbf{F}^{\prime}$. Specifically, $\textbf{F}^{\prime}$ combines $\textbf{F}$ with supplementary features $\textbf{F}_{sd}$ for the interaction decoder, while retaining the original features $\textbf{F}$ for the instance decoder.

\subsection{Multi-Modal Prompt-based Predictor}
\label{sec:CLIP}
\textbf{Motivation.} As discussed in the introduction, existing language-guided-only HOI detectors face challenges in generalization when dealing with entirely new categories that lack any relevant text
descriptions in the training data. They also struggle to address complex scenarios with high interaction ambiguity.
Therefore, we are motivated to incorporate visual prompts to enhance the effectiveness and generalization capability in open-world scenarios. 
Visual prompts can act as exemplars, allowing the model to detect the same objects/interactions, and they also aid in reducing semantic ambiguity present in textual prompts. Simultaneously, textual prompts provide semantic context to enhance the understanding of visual prompts. Given that CLIP~\cite{radford2021learning} is trained to align visual and textual representations through hundreds of millions of image-text pairs. we naturally leverage its pre-trained image and text encoders to encode both textual and visual prompts.

\myPara{Encoding Multi-Modal Prompts.}
\textbf{\textit{For textual prompt inputs,}} we can parse them into instance descriptions like "A photo of a [object]," and interaction descriptions like "A photo of a [man/child/...] [verb-ing] a [object]." These sentences can then be encoded to obtain object category embeddings $\boldsymbol{T}_{text}^{o}$ or interaction category embeddings $\boldsymbol{T}_{text}^{i}$.
\textbf{\textit{For visual prompt inputs,}} we can crop them into corresponding object ROI images and HOI ROI images using user-provided bounding boxes. These images can also be used to obtain object category embeddings $\boldsymbol{T}_{img}^{o}$ or interaction category embeddings $\boldsymbol{T}_{img}^{i}$.

\myPara{Multi-Modal Prompt-based Classifiers.} 
Formally, the final object category distributions $\boldsymbol{P}_o$ and HOI category distributions $\boldsymbol{P}_i$ can be calculated as,
\begin{gather}
    \boldsymbol{P}_o = \mathrm{softmax}(\boldsymbol{Q}^{\prime}_{o} * {\boldsymbol{T}_{text/img}^o}^{\intercal} ) \\
        \boldsymbol{P}_i = \mathrm{softmax}(\boldsymbol{Q}^{\prime}_{i} * {\boldsymbol{T}_{text/img}^i}^{\intercal})
\end{gather}
where $\boldsymbol{Q}^{\prime}_{o} \in \mathbb{R}^{ N \times C }$ and $\boldsymbol{Q}^{\prime}_{i} \in \mathbb{R}^{ N \times C }$ denotes object queries and interaction queries, respectively, and $\mathrm{softmax}$ indicates the row-wise softmax operation.

\subsubsection{Reformulate the Optimization of HOI}
\label{sec:train}
\textbf{Cross-modal Contrastive Learning.} \emph{MP-HOI} reformulate the HOI task as a similarity learning process between multi-modal prompts and objects/interactions via a unified contrastive loss. Thus, it can perform multi-modal prompt-to-object/interaction alignments for large-scale image-prompt pairs in \emph{Magic-HOI} and \emph{SynHOI}. In general, given $B$ image-prompt pairs $\{(\textbf{x}_{m},{\rm{prompt}}_{m})\}_{m=1}^{B}$ in our datasets, the unified contrastive loss $\mathcal{L}_{c}$ can be formulated as, 
\begin{equation}
    \mathcal{L}_{c} = -\frac{1}{B}\sum_{m=1}^{B}{\rm{log}}(\frac{{\rm{exp}}(S(\textbf{x}_{m},{\rm{prompt}}_{m})/{\tau})}{{\sum_{n=1}^{B}}{\rm{exp}}(S(\textbf{x}_{m},{\rm{prompt}}_{n})/{\tau})})
\end{equation}
where $S(\textbf{x}_{m},{\rm{prompt}}_{n})$ is prompt-to-image similarity between $m$-th image $\textbf{x}_{m}$ and $n$-th prompt ${\rm{prompt}}_{n}$. $\tau$ is a temperature to scale the logits.  Through $\mathcal{L}_{c}$, we could obtain the object contrastive
loss $\mathcal{L}_{c}^{o}$
and the interaction contrastive loss $\mathcal{L}_{c}^{i}$, which enable the model to learn generalizable and transferable objects/interactions representations from the large-scale categories.

\myPara{Overall Loss}. Following the query-based methods~\cite{liao2022gen,zhang2021mining}, we employ the Hungarian algorithm to match predictions to each ground-truth. The overall loss is computed between the matched predictions and their corresponding ground-truths, which includes the box regression loss $\mathcal{L}_{b}$, the intersection-over-union loss $\mathcal{L}_{g}$, the object contrastive
loss $\mathcal{L}_{c}^{o}$
, and the interaction contrastive loss $\mathcal{L}_{c}^{i}$,

\begin{equation}
    \mathcal{L} = \lambda_{b}\mathcal{L}_{b} +  \lambda_{g}\mathcal{L}_{g} + \lambda_{c}^{o} \mathcal{L}_{c}^{o} + \lambda_{c}^{i}\mathcal{L}_{c}^{i},
\end{equation}
where $\mathcal{L}_{b}$ and $\mathcal{L}_{g}$ contain both human and object localization. $\lambda_{b}$, $\lambda_{g}$, $\lambda_{c}^{o}$ and $\lambda_{c}^{i}$ are used to adjust the weights of each loss component.

\subsection{Training and Inference Details}

\textbf{Training Details.} With the unified cross-modality contrastive loss that we have defined, we are able to jointly optimize multi-modal prompts. Considering the training cost, we employ a $50\%$ probability to randomly select either a visual prompt or a textual prompt for each iteration. To use visual prompts for training, we sample two images containing a person paired with the exact same single/multiple objects and single/multiple interaction categories from the dataset, and then choose one of them as the visual prompt.

\myPara{Inference Details.} 
Users can freely choose from available prompts. We present quantitative results based on textual prompts to ensure a fair comparison.

\section{Experiments}
\subsection{Experimental Settings}
\begin{table}[t]
\begin{center}
\resizebox{1.\linewidth}{!}{
  \begin{threeparttable}
\begin{tabular}{llccc}
\hline
\multirow{3}{*}{Method} & \multirow{3}{*}{Backbone} & \multicolumn{3}{c}{Default} \\
\cmidrule{3-5} 
&& Full & Rare & Non-Rare  \\
\hline\hline
\rowcolor{Gray!80} \multicolumn{5}{l}{\emph{Expert Models}} \\
QPIC \cite{tamura2021qpic}  & ResNet-50 &  29.07 & 21.85&  31.23 \\
CDN \cite{zhang2021mining}  & ResNet-50 & 31.44& 27.39 &32.64  \\
DOQ \cite{qu2022distillation} & ResNet-50 & 33.28 & 29.19 & 34.50  \\
IF \cite{liu2022interactiveness} & ResNet-50 & 33.51 & 30.30 & 34.46 \\
GEN-VLKT \cite{liao2022gen} & ResNet-50 & 33.75	& 29.25& 	35.10\\
PViC~\cite{zhang2023exploring} & ResNet-50 & \underline{33.80} & \underline{29.28} & \underline{35.15} \\
QAHOI~\cite{chen2021qahoi} & Swin-L& 35.78 & 29.80 & 37.56 \\
FGAHOI~\cite{ma2023fgahoi} & Swin-L& 37.18 & 30.71 & 39.11   \\
PViC~\cite{zhang2023exploring} & Swin-L & \underline{43.35} & \underline{42.25} &  \underline{43.69} \\
\rowcolor{Gray!80} \multicolumn{5}{l}{\emph{Pre-trained Models}} \\
RLIPv1~\cite{yuan2022rlip} & ResNet-50 &  13.92 & 11.20  & 14.73 \\
RLIPv1~\cite{yuan2022rlip}$^{\dagger}$ & ResNet-50 &  30.70 & 24.67 & 32.50 \\
RLIPv2~\cite{yuan2023rlipv2} & ResNet-50 & 17.79 & 19.64  & 17.24 \\
RLIPv2~\cite{yuan2023rlipv2}$^{\dagger}$ & ResNet-50 & 35.38 & 29.61 &  37.10 \\
\rowcolor{Gray!80} \multicolumn{5}{l}{\emph{Generalist Models}} \\
\emph{MP-HOI}-S                  & ResNet-50  & 36.50{\color{Red}$\uparrow_{2.70}$}  & 35.48{\color{Red}$\uparrow_{6.20}$} & 36.80{\color{Red}$\uparrow_{1.65}$}   \\  
\emph{MP-HOI}-L                  & Swin-L &  \textbf{44.53}{\color{Red}$\uparrow_{1.18}$}  & \textbf{44.48}{\color{Red}$\uparrow_{2.23}$} & \textbf{44.55}{\color{Red}$\uparrow_{0.86}$}     \\     
\hline
\end{tabular}
		  \begin{tablenotes}   
        \footnotesize               
\item[1] RLIP V2: VG+COCO+O365, $2200K+$ pretraining data.
\item[2] \emph{MP-HOI}: \emph{Magic-HOI}+\emph{SynHOI}, $286K+$ training data.
      \end{tablenotes}    
      \end{threeparttable}       
}
\end{center}
\vspace{-0.6cm}
\caption{Performance comparison on HICO-DET in terms of mAP (\%). $\dagger$ indicates the models are fully fine-tuned on HICO-DET. 
The results highlighted in \underline{underlined} represent state-of-the-art performance among expert models, which we primarily compare to. 
} 
\label{tab:HICO-DET}
\end{table}

\textbf{Datasets and Evaluation Metrics.} We evaluate our models on four benchmarks: HICO-DET~\cite{chao2018learning}, V-COCO~\cite{gupta2015visual}, SwiG~\cite{wang2021discovering} and HCVRD~\cite{zhuang2018hcvrd}. The mean Average Precision (mAP) is used as the evaluation metric, following standard protocols~\cite{zhang2021mining, liao2022gen}. 

\noindent \textbf{Implementation Details.}
We implement two variant architectures of \emph{MP-HOI}: \emph{MP-HOI}-S, and \emph{MP-HOI}-L, where ‘S’ and
‘L’ refer to small and large, respectively. For \emph{MP-HOI}-S, we use ResNet-50 as the backbone and a six-layer vanilla Transformer encoder~\cite{carion2020end} as the feature extractor. Both the human-object decoder and interaction decoder are three-layer vanilla Transformer decoders. For \emph{MP-HOI}-L, we employ Swin-L~\cite{liu2021swin} as the backbone, with Transformer decoders of six layers. 
The CLIP-based model and diffusion model are frozen during training for all the settings. 
For a fair comparison with previous studies, we employ the ViT-B variant of the CLIP model within the \emph{MP-HOI}-S. Conversely, to enhance performance in real-world scenarios, the ViT-L variant is utilized within the \emph{MP-HOI}-L.

\begin{table}[t]
\begin{center}
\resizebox{.8\linewidth}{!}{
\begin{tabular}{llccc}
\hline
\multirow{3}{*}{Method} & \multirow{3}{*}{Backbone} & \multicolumn{3}{c}{Default} \\
\cmidrule{3-5} 
&& Full & Rare & Non-Rare  \\
\hline\hline
\rowcolor{Gray!80} \multicolumn{5}{l}{\emph{Expert Models}} \\
GEN-VLKT~\cite{liao2022gen} & ResNet-50 & 10.87 & 10.41 & 20.91	\\
\rowcolor{Gray!80} \multicolumn{5}{l}{\emph{Generalist Models}} \\
\emph{MP-HOI}-S                  & ResNet-50  & 12.61 & 14.78 & 20.28  \\  
\emph{MP-HOI}-L                  & Swin-L & \textbf{16.21} & \textbf{18.59} & \textbf{25.76}  \\
\hline
\end{tabular}}
\end{center}
\vspace{-1.4em}
\caption{Performance comparison on SWiG in terms of mAP (\%). 
Given that GEN-VLKT is one of the most powerful and representative expert models, we train it on SWiG for comparison.
} 
\label{tab:SWiG}
\vspace{-1.2em}
\end{table}

\begin{table}[t]
\begin{center}
\resizebox{.8\linewidth}{!}{
\begin{tabular}{llccc}
\hline
\multirow{3}{*}{Method} & \multirow{3}{*}{Backbone} & \multicolumn{3}{c}{Default} \\
\cmidrule{3-5} 
&& Full & Rare & Non-Rare  \\
\hline\hline
\rowcolor{Gray!80} \multicolumn{5}{l}{\emph{Expert Models}} \\
GEN-VLKT~\cite{liao2022gen} & ResNet-50 & 6.58 & 5.81 & 14.02 	\\
\rowcolor{Gray!80} \multicolumn{5}{l}{\emph{Generalist Models}} \\
\emph{MP-HOI}-S                  & ResNet-50   & 8.08 & 6.86 & 14.31   \\  
\emph{MP-HOI}-L                  & Swin-L & \textbf{11.29} & \textbf{9.01} & \textbf{18.68}   \\ \hline
\end{tabular}}
\end{center}
\vspace{-0.6cm}
\caption{Performance comparison on HCVRD in terms of mAP (\%).
We train GEN-VLKT on HCVRD for comparison.
} 
\label{tab:HCVRD}
\end{table}

\begin{table}[t]
\begin{center}
\resizebox{1.\linewidth}{!}{
\begin{tabular}{llcc}
\hline
Method & Backbone & AP (Scenario 1)  & AP (Scenario 2) \\ \hline\hline
\rowcolor{Gray!80} \multicolumn{4}{l}{\emph{Expert Models}} \\
IDN \cite{li2020hoi} &  ResNet-50 &  53.3 & 60.3\\
FCL \cite{hou2021detecting} &  ResNet-50 & 52.4 & - \\
SCG \cite{zhang2021spatially}  &  ResNet-50 &  54.2 & 60.9 \\
HOTR~\cite{kim2021hotr} &  ResNet-50  & 55.2 & 64.4 \\
QPIC~\cite{tamura2021qpic} &  ResNet-50 & 58.8 & 61.0 \\
CDN \cite{zhang2021mining}  &  ResNet-50 & 61.7 & 63.8 \\
GEN-VLKT~\cite{liao2022gen}&  ResNet-50 &  62.4	& 64.5 \\ 
\hline  
\rowcolor{Gray!80} \multicolumn{4}{l}{\emph{Generalist Models}} \\
\rowcolor{blue!5} \multicolumn{4}{l}{\emph{Zero-shot Testing}} \\
\emph{MP-HOI}-S & ResNet-50  & 37.5 & 44.2 \\   
\rowcolor{blue!5} \multicolumn{4}{l}{\emph{10\% Training Data}} \\
\emph{MP-HOI}-S & ResNet-50   & 57.7 & 60.2 \\ 
\rowcolor{blue!5} \multicolumn{4}{l}{\emph{100\% Training Data}} \\
\emph{MP-HOI}-S & ResNet-50   & \textbf{66.2} & \textbf{67.6} \\ 
\hline  
\end{tabular}
}
\end{center}
\vspace{-0.5cm}
\caption{Performance comparison on V-COCO.}
\label{tab:v-coco}
\vspace{-0.6cm}
\end{table}

\subsection{General HOI Detection}
\emph{MP-HOI} is a generalist HOI detector trained on the unified dataset-\emph{Magic-HOI}. Since \emph{Magic-HOI} unifies multiple datasets, we could directly evaluate \emph{MP-HOI} on these benchmarks. We compare its performance with two types of models: expert models and pretrained models. As there are no standard results available like mAP on SWiG-HOI and HCVRD datasets for comparison, we train the GEN-VLKT~\cite{liao2022gen}, which is one of the representative expert models, as the baseline for our comparison.

\myPara{HICO-DET.} Compared with the expert models, \emph{MP-HOI} outperforms them by a significant margin under both ResNet-50 and Swin-L backbones, notably achieving a remarkable $6.20$ improvement in mAP in the rare setting. 
Compared with the pre-trained model, \emph{MP-HOI} also surpasses RLIPv2~\cite{yuan2023rlipv2}, which is pretrained on over $2200K$ data and fine-tuned on HICO-DET. Finally, \emph{MP-HOI}-L achieves a new SOTA performance with $44.53$ mAP.

\myPara{SWiG-HOI \& HCVRD.} Compared with GEN-VLKT~\cite{liao2022gen}, \emph{MP-HOI} exhibits superior performance on these two datasets that contain more HOI concepts. This highlights \emph{MP-HOI}'s capability to serve as a generalist HOI detector.

\begin{table*}[t]
    \begin{center}
                            \begin{minipage}{0.3\linewidth}
    
        \resizebox{\linewidth}{!}{
            \makeatletter\def\@captype{table}\makeatother
\begin{tabular}{cc|ccc}
\hline
$\boldsymbol{F}_{sd}$ & $\boldsymbol{F}_{clip}^{img}$ &   Full & Rare & Non-rare                \\ \hline
& & 31.99 & 29.63 &  32.70                           \\
  \checkmark & &  32.92  & \textbf{31.29} & 33.41                                  \\
  \checkmark  &  \checkmark &     \textbf{34.41} & 31.07 & \textbf{35.40}                          \\
\hline
\end{tabular}
            }
                            \vspace{-0.38cm}
\caption{{Ablation results of two key representations on HICO-DET.}}
            \label{tab:representations}
            \vspace{-0.1cm}
        \end{minipage}
            \quad
\begin{minipage}{0.3\linewidth}
        \resizebox{\linewidth}{!}{
            \makeatletter\def\@captype{table}\makeatother
\begin{tabular}{c|ccc}
\hline
Time step                   & Full         & Rare           & Non-Rare           \\
\hline
0                      & \textbf{34.41} & \textbf{31.07} & \textbf{35.40}      \\
100                   & 34.03 & 30.58 & 35.02              \\
500                    & 33.59 & 29.80 & 34.71            \\
\hline
\end{tabular}
            }
                     \vspace{-0.38cm}
\caption{{Ablation results of different diffusion time steps on HICO-DET.}}
            \label{tab:time-step}
            \vspace{-0.25cm}
        \end{minipage}
  \quad
                        \begin{minipage}{0.35\linewidth}
        \resizebox{\linewidth}{!}{
            \makeatletter\def\@captype{table}\makeatother
\begin{tabular}{cc|ccc}
\hline
Adaptor$\alpha$   & Adaptor$\beta$                  & Full         & Rare           & Non-Rare           \\
\hline
\checkmark   &                   & 	34.08 &	30.64	& 35.08 \\
 & \checkmark                   & 33.76	&29.89	& 34.82              \\
\checkmark    & \checkmark                  & \textbf{34.41} & \textbf{31.07} & \textbf{35.40}       \\
\hline
\end{tabular}
            }
              \vspace{-0.25cm}
\caption{{Ablation results of two scene adaptors on HICO-DET.}}
            \label{tab:adaptors}
            \vspace{-0.25cm}
        \end{minipage}
    \end{center}
    \vspace{-0.3cm}

\end{table*}

\begin{table*}[t]
    \begin{center}
                \begin{minipage}{0.30\linewidth}
        \resizebox{\linewidth}{!}{
            \makeatletter\def\@captype{table}\makeatother
{\tiny \begin{tabular}{cc|ccc}
\hline
$\mathcal{L}_{c}^{o}$ & $\mathcal{L}_{c}^{i}$                  & Full         & Rare           & Non-Rare           \\
\hline  

 & & 34.41 & 31.07 & 35.40             \\
\checkmark   & & 34.56 & 31.29 & 35.43  \\
\checkmark   & \checkmark & \textbf{34.82} & \textbf{31.87}& \textbf{35.49} \\

\hline
\end{tabular}}
}
                     \vspace{-0.2cm}
\caption{{The ablation study of contrastive losses at two levels on HICO-DET.}}
            \label{tab:contrastive}
            \vspace{-0.3cm}
        \end{minipage}
  \quad
                        \begin{minipage}{0.34\linewidth}
        \resizebox{\linewidth}{!}{
            \makeatletter\def\@captype{table}\makeatother
{\tiny \begin{tabular}{cc|ccc}
\hline
\textbf{T.} & \textbf{V.}                 & Full         & Rare           & Non-Rare           \\
\hline  \checkmark   & & 34.82 & 31.87& 35.49\\
\checkmark   & \checkmark & \textbf{35.18} & \textbf{32.16}& \textbf{35.57} \\

\hline
\end{tabular}}
}
                     \vspace{-0.3cm}
\caption{{The ablation study of multi-modal prompts on HICO-DET. $\textbf{T.}$ and $\textbf{V.}$ denotes textual prompt and visual prompt, respectively.}}
            \label{tab:multi-modal ablation}
            \vspace{-0.3cm}
        \end{minipage}
  \quad
                        \begin{minipage}{0.31\linewidth}
                        \vspace{-1em}
        \resizebox{\linewidth}{!}{
            \makeatletter\def\@captype{table}\makeatother
\begin{tabular}{c|ccc}
\hline
       Training Data          & Full         & Rare           & Non-Rare   \\ \hline 
    HICO-DET    & 35.18 & 32.16& 35.57  \\
       \emph{Magic-HOI}    & 35.93 & 34.68 & 36.26  \\
       +\emph{SynHOI} & \textbf{36.50} & \textbf{35.48} & \textbf{36.80}    \\
\hline 
\end{tabular}
            }
              \vspace{-0.3cm}
\caption{{The impact of the scale of training data.}}
            \label{tab:data_scale}
            \vspace{-0.3cm}
        \end{minipage}
    \end{center}
    \vspace{-0.3cm}

\end{table*}

\begin{figure*}[h!]
    \centering
    \begin{subfigure}{.19\textwidth}
        \centering
        \includegraphics[height=2cm]{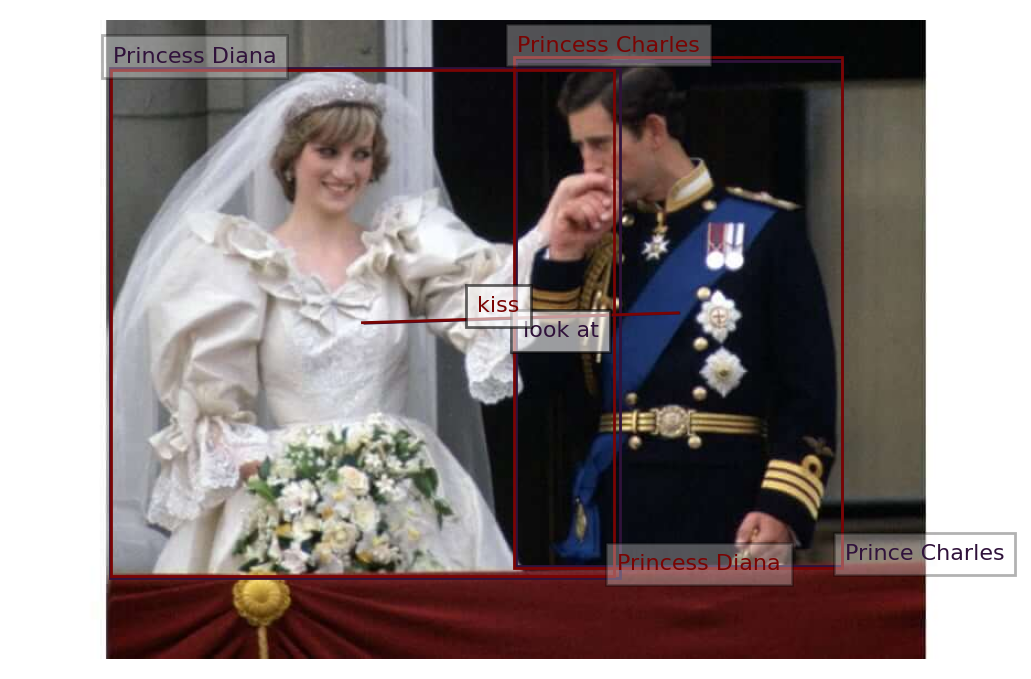}
        \caption{}
    \end{subfigure}%
    \begin{subfigure}{.19\textwidth}
        \centering
        \includegraphics[height=1.8cm]{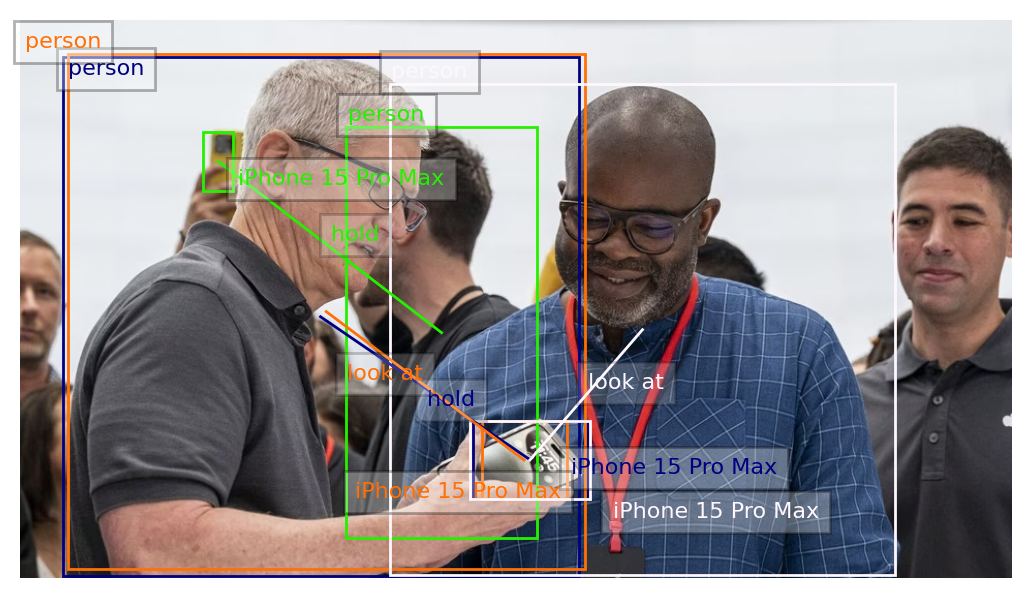}
        \caption{}
    \end{subfigure}
    \begin{subfigure}{.19\textwidth}
        \centering
        \includegraphics[height=2cm]{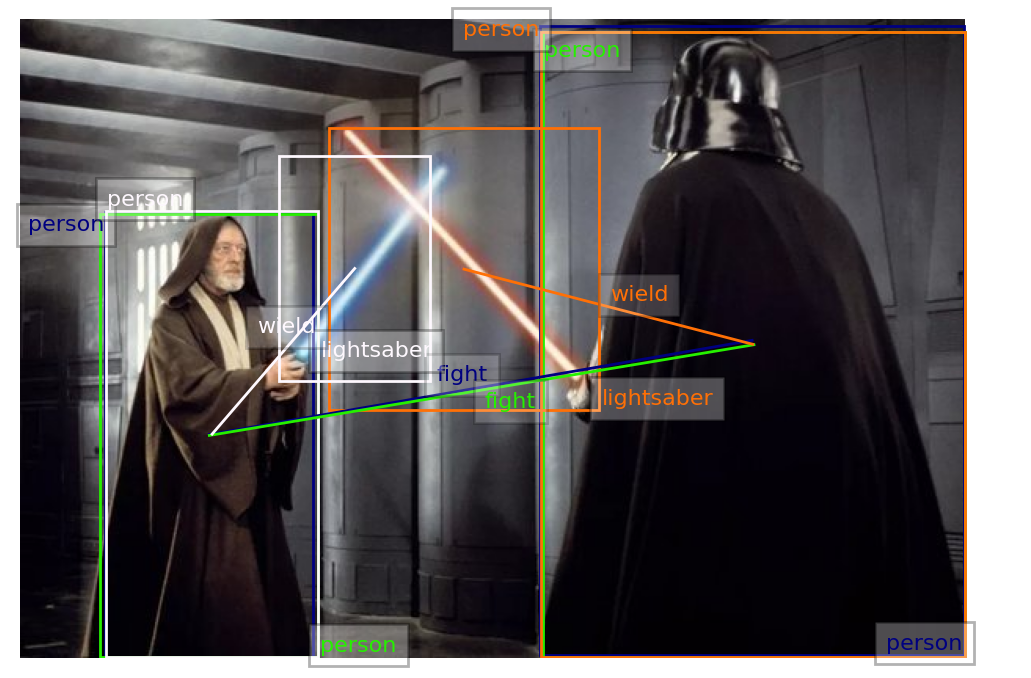}
        \caption{}
    \end{subfigure}
    \begin{subfigure}{.19\textwidth}
        \centering
        \includegraphics[height=1.8cm]{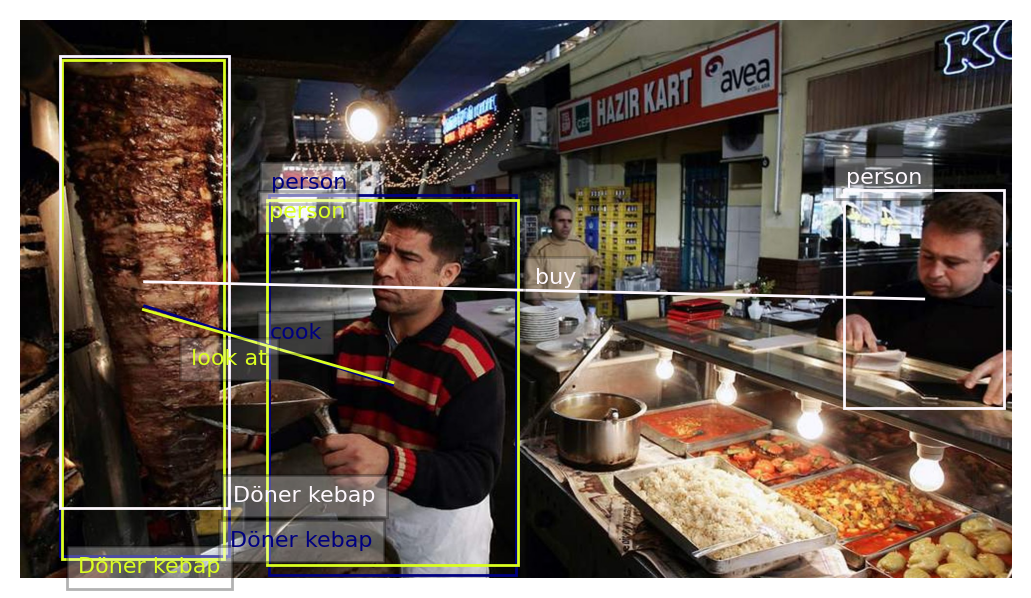}
        \caption{}
    \end{subfigure}
    \begin{subfigure}{.19\textwidth}
        \centering
        \includegraphics[height=2cm]{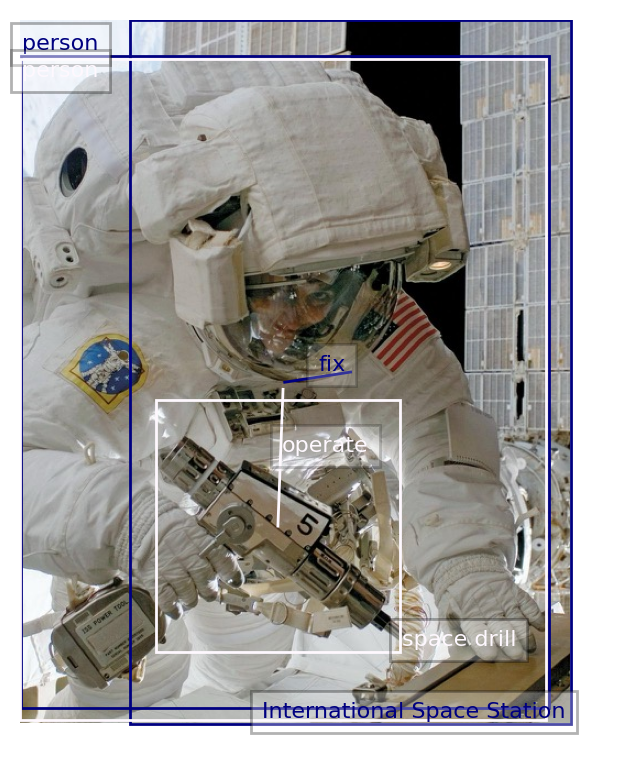}
        \vspace{-0.1em}\caption{}
    \end{subfigure}
    \vspace{-0.30cm}
    \caption{In-the-wild test based on arbitrary textual prompts. Each HOI triplet is represented in the same color.}
    \vspace{-0.2cm}
    \label{fig:wild_demo}
\end{figure*}

\begin{figure*}[h!]
    \centering    
    \begin{subfigure}{.19\textwidth}
        \centering
        \includegraphics[height=2cm]{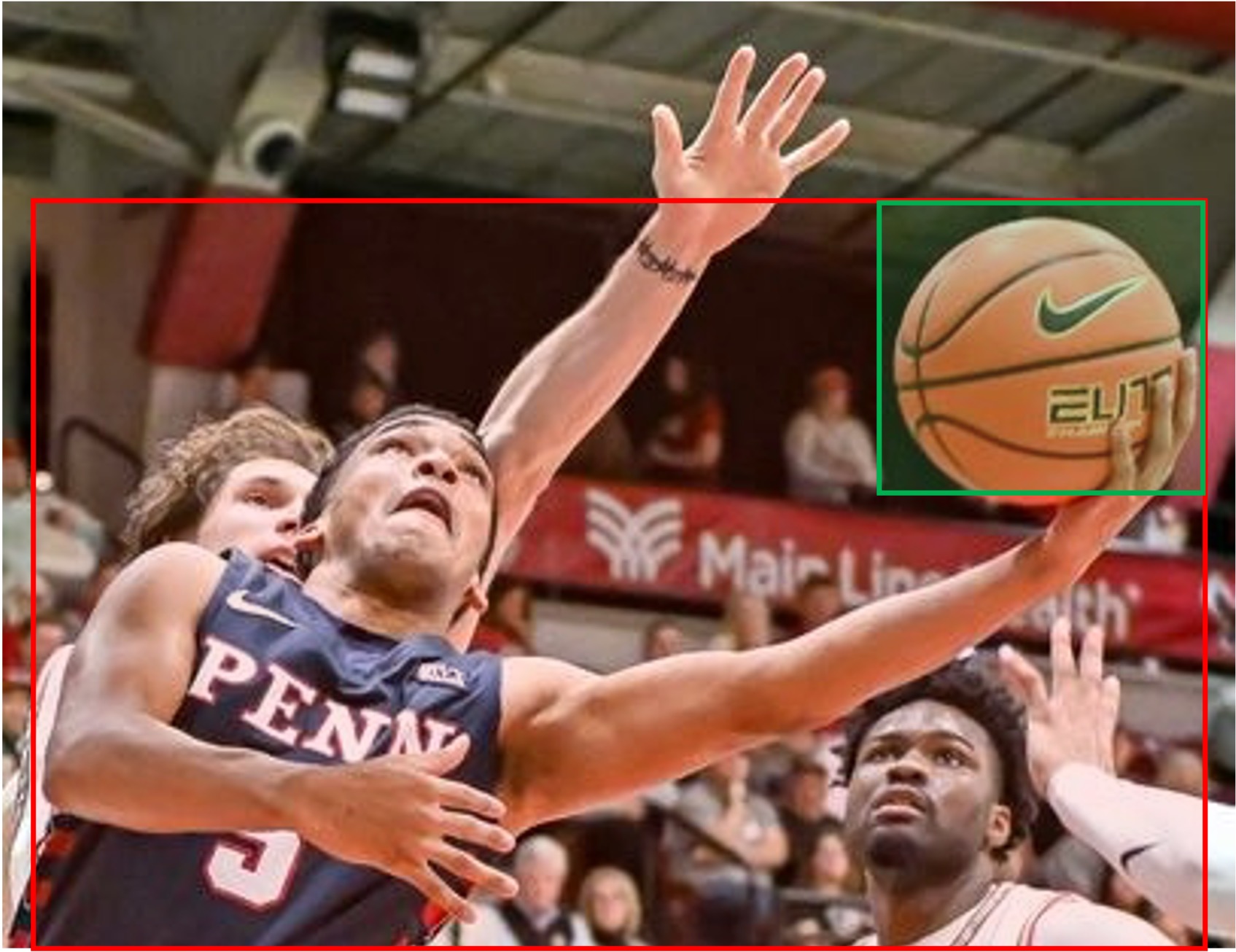}
    \end{subfigure}
    \begin{subfigure}{.19\textwidth}
        \centering
        \includegraphics[height=2cm]{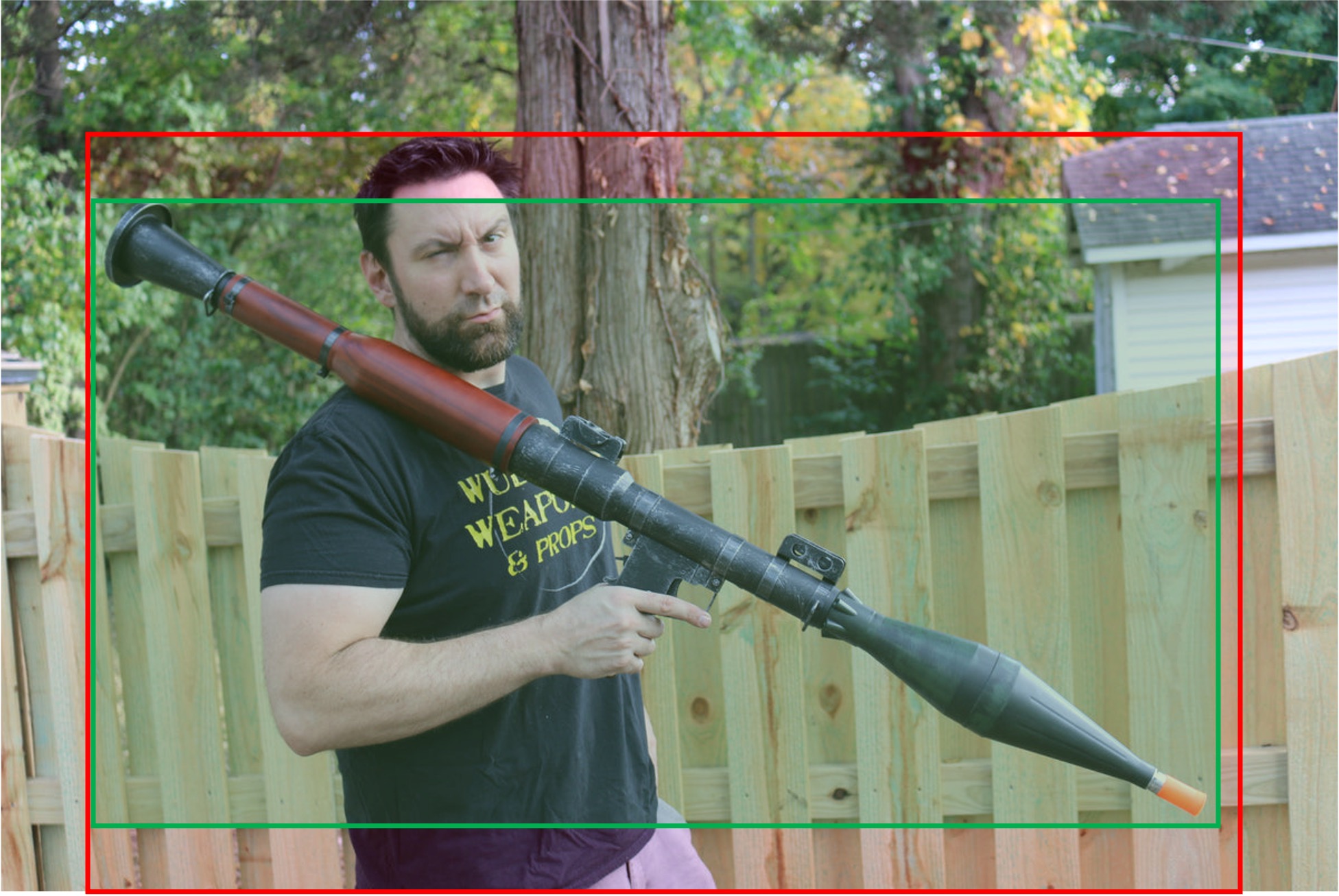}
    \end{subfigure}
    \begin{subfigure}{.19\textwidth}
        \centering
        \includegraphics[height=2cm]{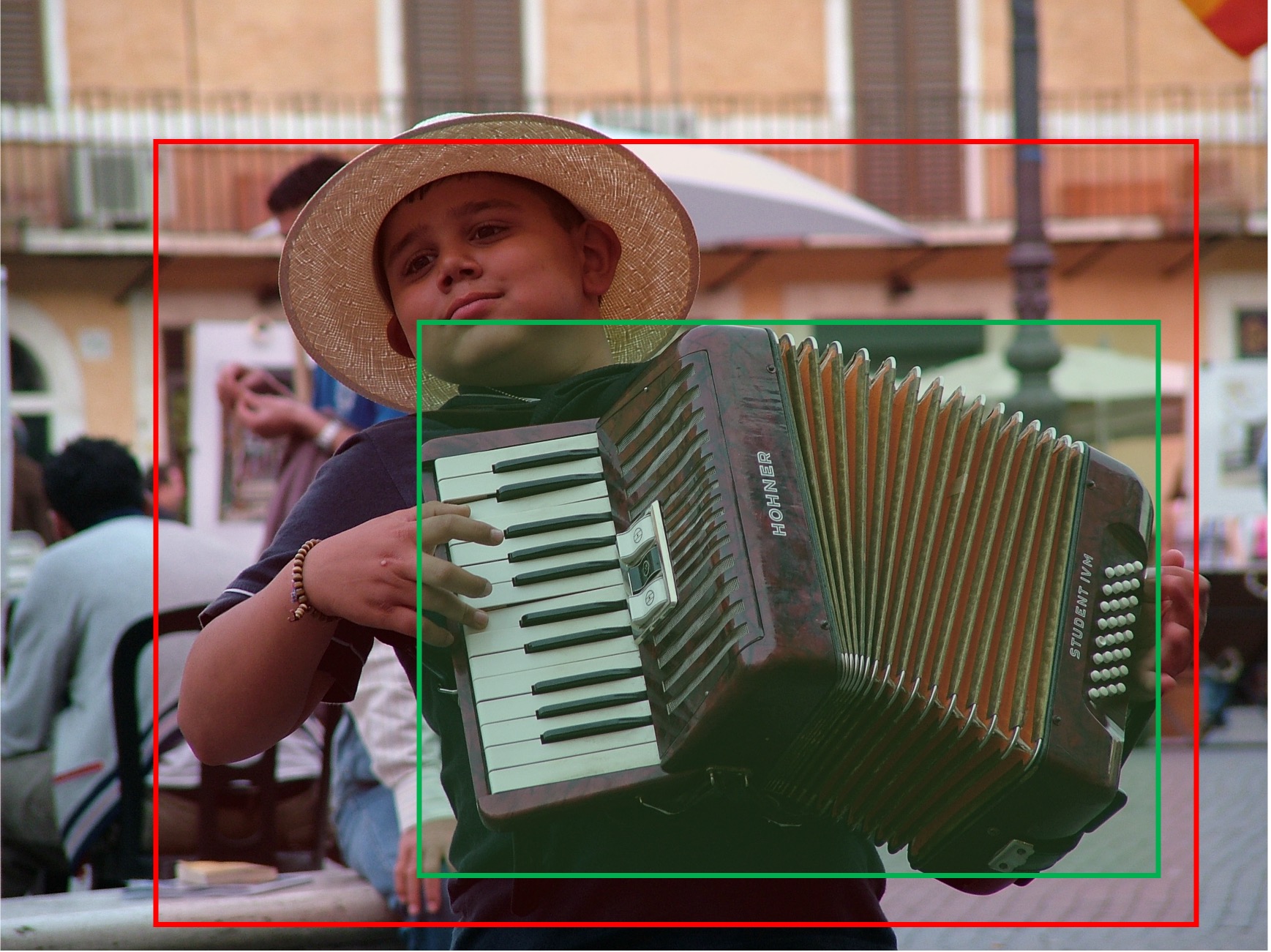}
    \end{subfigure}
    \begin{subfigure}{.19\textwidth}
        \centering
        \includegraphics[height=2cm]{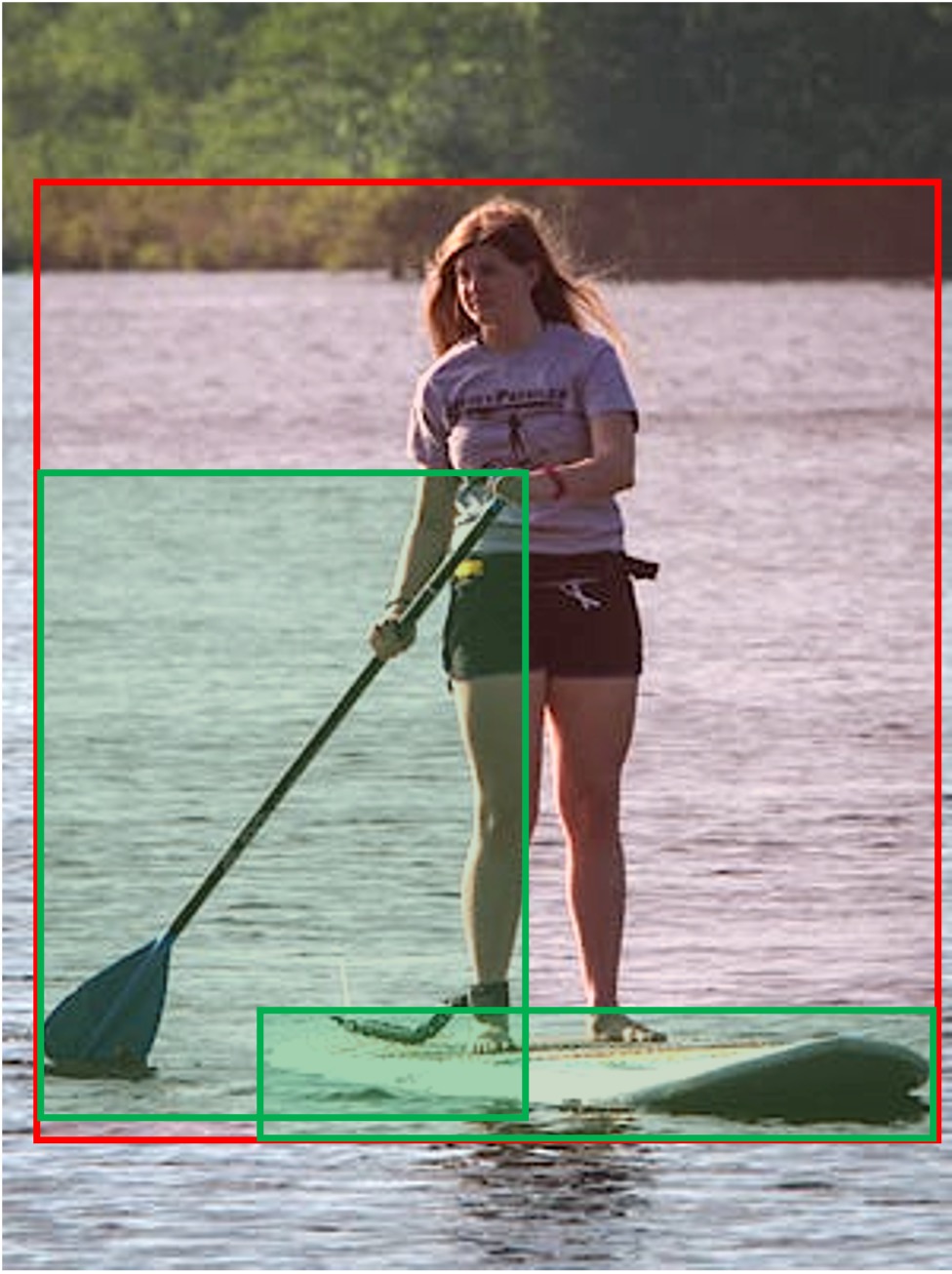}
    \end{subfigure}%
    \begin{subfigure}{.19\textwidth}
        \centering
        \includegraphics[height=2cm]{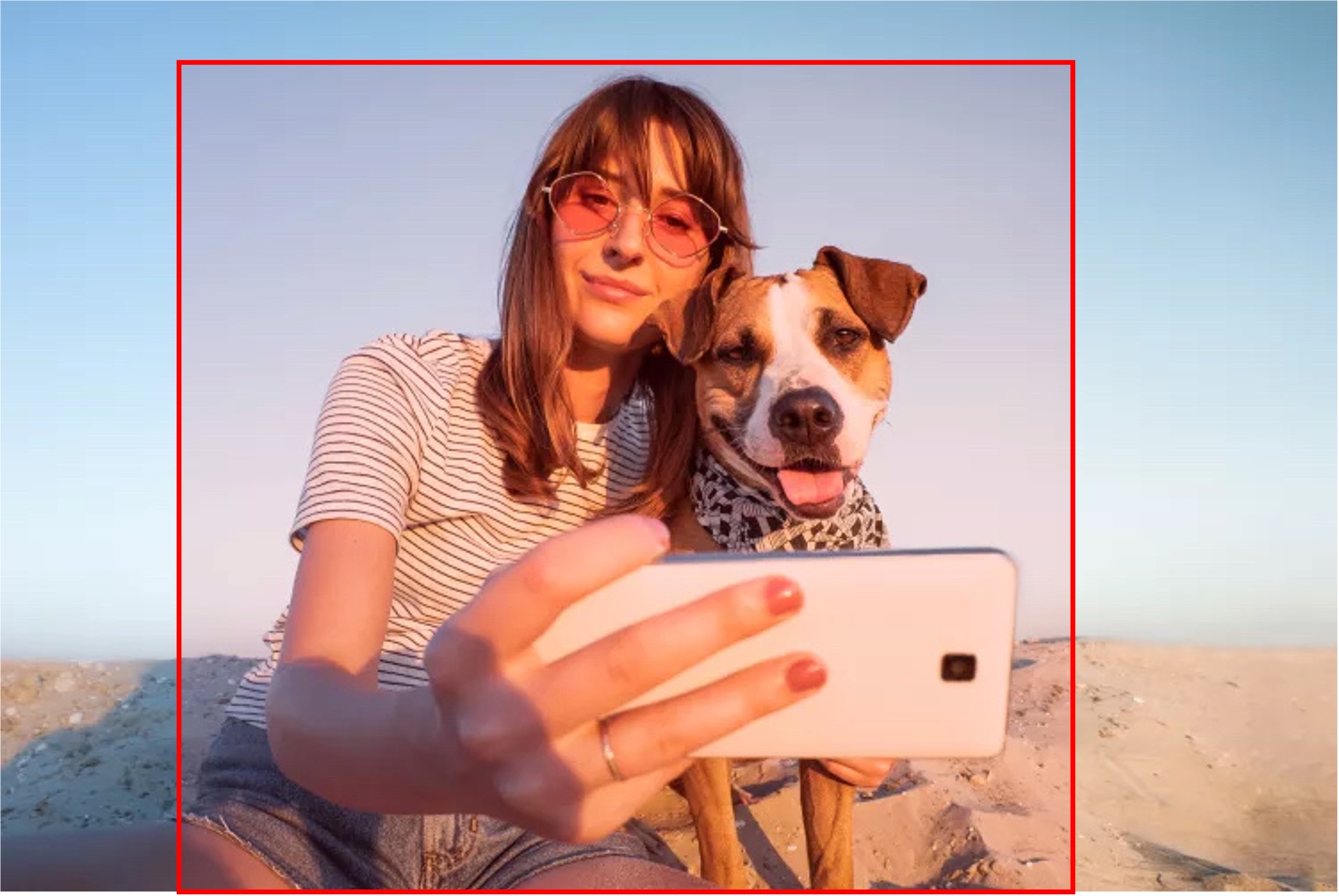}
    \end{subfigure}

    \begin{subfigure}{.19\textwidth}
        \centering
        \includegraphics[height=2cm]{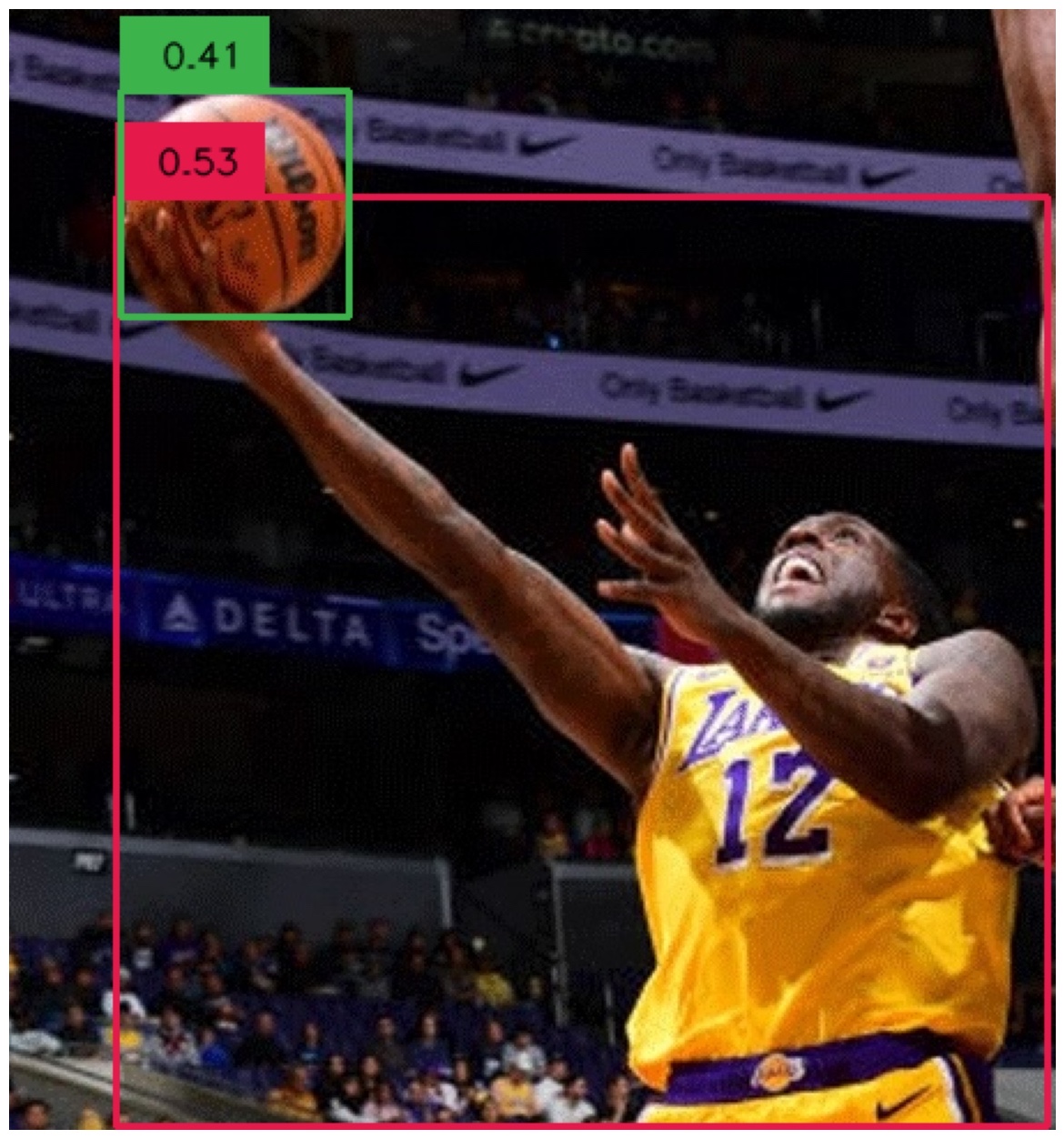}
        \caption{}
    \end{subfigure}
    \begin{subfigure}{.19\textwidth}
        \centering
        \includegraphics[height=2cm]{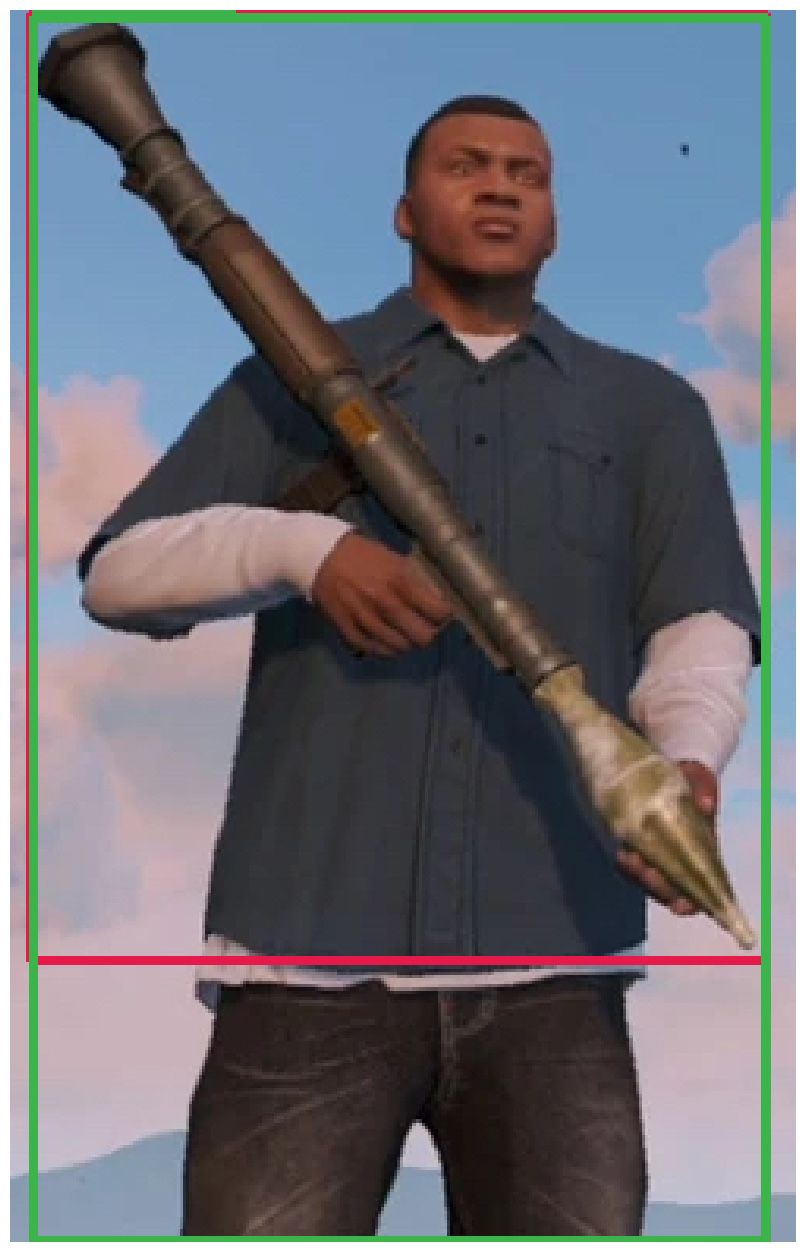}
        \caption{}
    \end{subfigure}
    \begin{subfigure}{.19\textwidth}
        \centering
        \includegraphics[height=2cm]{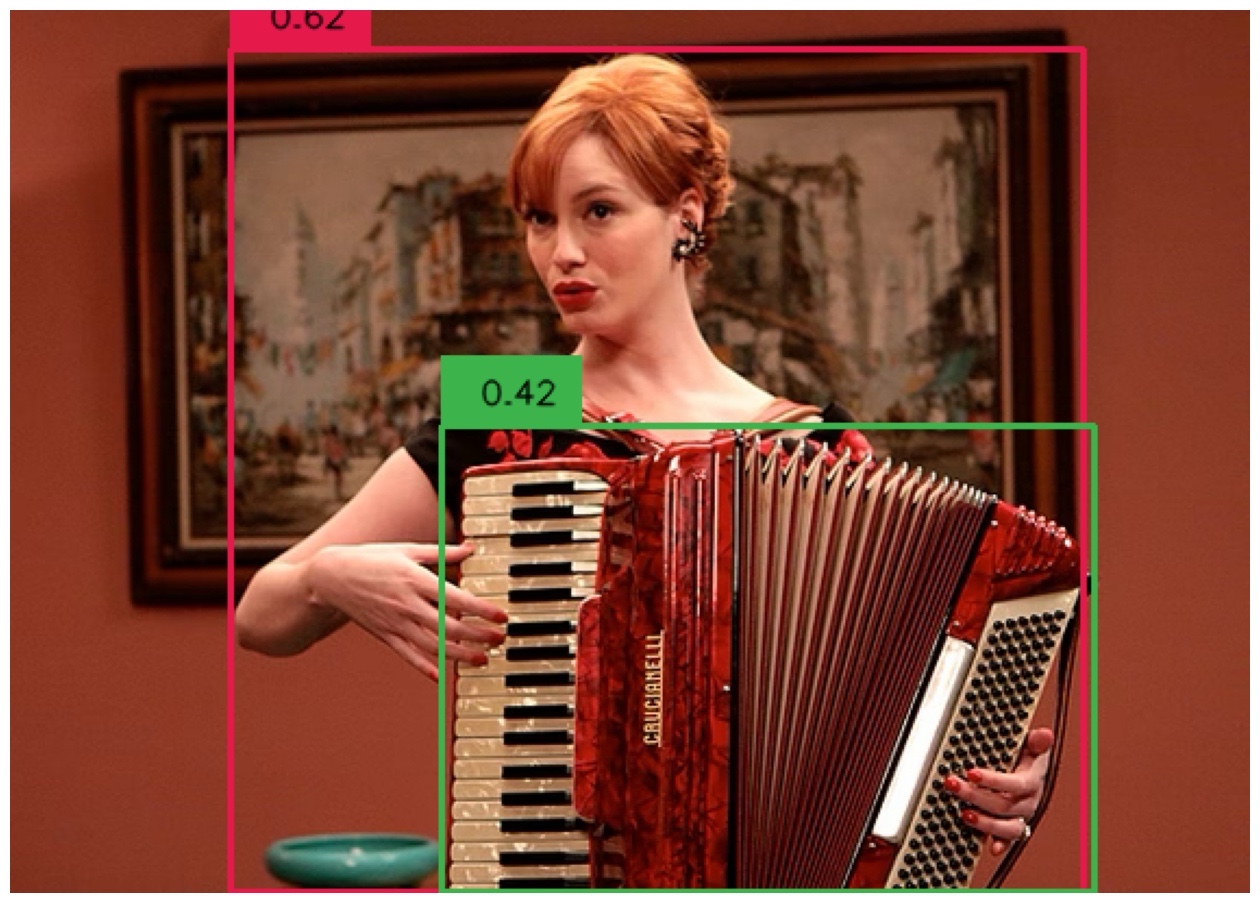}
        \caption{}
    \end{subfigure}
    \begin{subfigure}{.19\textwidth}
        \centering
        \includegraphics[height=2cm]{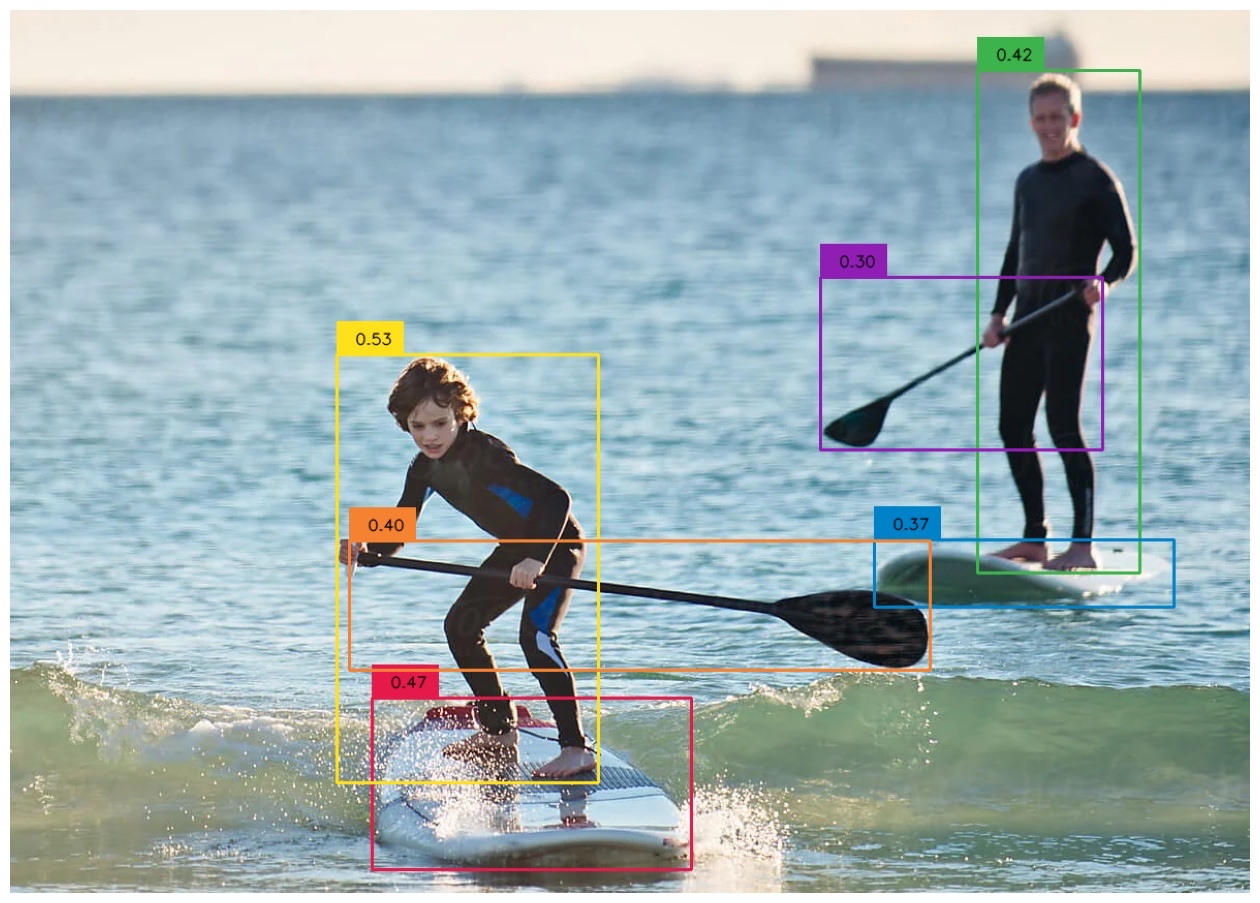}
        \caption{}
    \end{subfigure}%
    \begin{subfigure}{.19\textwidth}
        \centering
        \includegraphics[height=1.8cm]{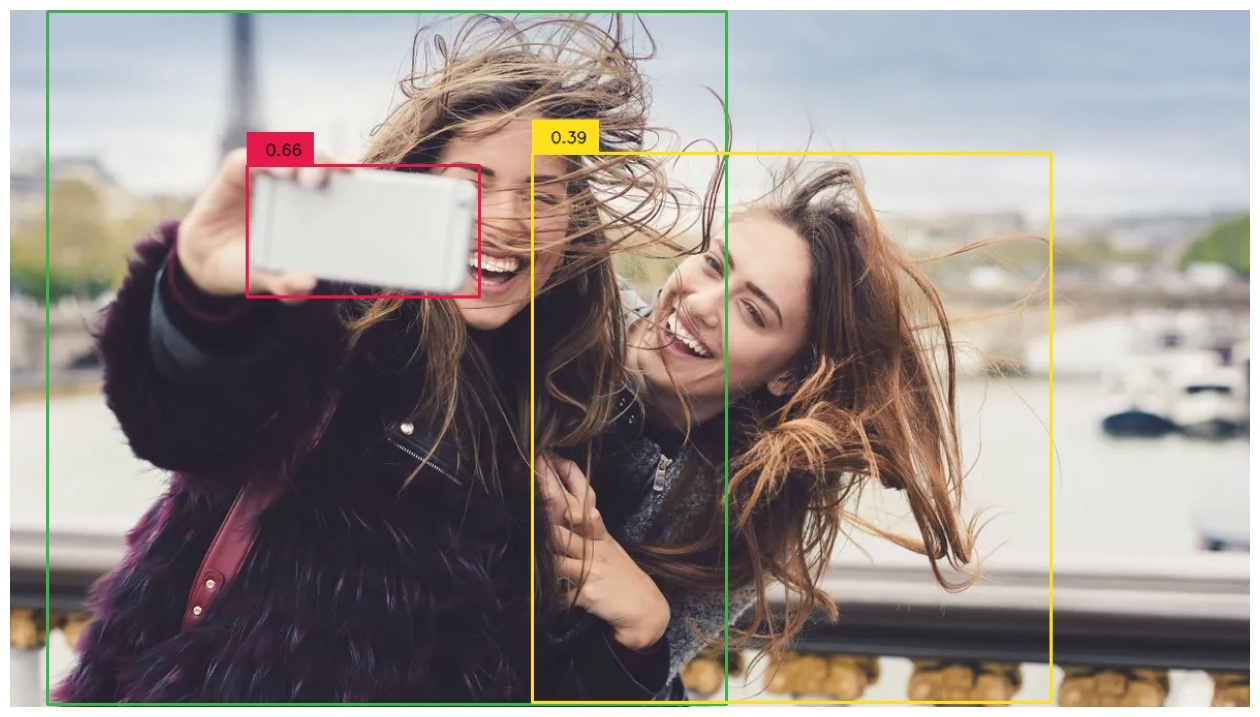}
        \caption{}
    \end{subfigure}
    \vspace{-0.35cm}
    \caption{In-the-wild test based on arbitrary visual prompts. The first row is the visual prompt (green boxes define objects and red boxes define interactions), and the second row is the corresponding detection results. We highlight the powerful HOI detection performance, the capability to detect single HOIs (a, b, c), as well as a specific HOI composition (d), and the flexibility to integrate the visual prompt for interaction definition and textual prompt for object definitions (e.g., woman and phone) for the test (e).}
    \label{fig:wild_visual_demo}
    \vspace{-0.45cm}
\end{figure*}

\subsection{Zero-Shot HOI Detection}
\vspace{-0.1cm}

We demonstrate the generalization of the model trained on our proposed \emph{Magic-HOI} and \emph{SynHOI} datasets by evaluating it on the unseen dataset-V-COCO. As shown in Tab.~\ref{tab:v-coco}, \emph{MP-HOI} exhibits excellent zero-shot performance on V-COCO. After fine-tuning the model on just $10\%$ of the training data, it achieves an impressive AP of $57.7$, surpassing some expert models that are fully trained on V-COCO. Moreover, when utilizing the full training data, \emph{MP-HOI} outperforms all expert models by a significant margin.

\vspace{-0.1cm}
\subsection{Ablation Study}
\vspace{-0.1cm}

\myPara{Two Key Representations.} As described in Sec.~\ref{sec:overview}, we introduce $\textbf{F}_{sd}$ from stable diffusion and $\textbf{F}_{clip}^{img}$ from CLIP image encoder, into \emph{MP-HOI}. Tab.~\ref{tab:representations} demonstrates their effectiveness. Initially, we establish a baseline model following the design of GEN-VLKT~\cite{liao2022gen} while excluding the knowledge distillation component. Subsequently, the incorporation of $\textbf{F}_{sd}$ leads to significant improvements, particularly in terms of rare mAP. This suggests that the internal features in stable diffusion play a crucial role in enhancing the representation of rare categories in HICO-DET. Additionally, the integration of $\textbf{F}_{clip}^{img}$ further improves the non-rare AP, as it aligns the  queries with multi-modal prompts.

\myPara{Diffusion Time Steps.} We investigate the effectiveness of different diffusion steps in extracting $\textbf{F}_{sd}$, similarly to~\cite{baranchuk2021label,xu2023open}. Diffusion models control the noise distortion added to the input image by varying the value of $t$. Stable diffusion~\cite{rombach2022high} uses a total of $1000$ time steps. We set $t$ values to $0, 100, 500$ for ablation studies. As demonstrated in Tab.~\ref{tab:time-step}, the best performance is achieved when $t=0$.  It is worth noting that using $\textbf{F}_{sd}$ extracted from input images with higher noise levels would decrease performance and potentially impact the learning of interactions.

\myPara{Two Scene-Aware Adaptors.} Tab.~\ref{tab:adaptors} illustrates that both scene-aware adaptors effectively facilitate Stable Diffusion and the CLIP image encoder in extracting feature representations aligned with our \emph{MP-HOI} framework and HOI task, thereby improving performance.

\myPara{Contrastive Learning.} As in Sec.~\ref{sec:train}, we introduce the object contrastive
loss $\mathcal{L}_{c}^{o}$
and the interaction contrastive loss $\mathcal{L}_{c}^{i}$. We leverage these two losses to facilitate alignment between multi-modal prompts and objects/interactions. The results in Tab.~\ref{tab:contrastive} demonstrate the effectiveness of both loss items, particularly in improving performance on rare classes, where the model exhibits improved discrimination between objects and interactions.

\myPara{Multi-modal Prompts.} As shown in Tab.~\ref{tab:multi-modal ablation}, integrating
visual prompts during training could enhance performance compared to using only textual prompts. This improvement is due to the inherent semantic ambiguity in textual prompts, which visual prompts help to reduce.

\myPara{The Scale of Training Data.} Beyond solely training on HICO-DET, we incorporate our \emph{Magic-HOI} dataset that includes five additional datasets for training. As in Tab.~\ref{tab:data_scale}, this leads to improved performance and expands the range of object and interaction categories covered. Furthermore, the inclusion of our \emph{SynHOI} dataset could also effectively enhance performance, particularly on rare classes.

\subsection{HOI Detection in the Open World}
In addition to the zero-shot experiments, we also conduct open-world
testing on arbitrary images from the Internet as shown in Fig.~\ref{fig:wild_demo} and Fig.~\ref{fig:wild_visual_demo}. 
Our \emph{MP-HOI} could leverage arbitrary textual and visual prompts to detect HOIs. Interestingly, we find \emph{MP-HOI} could support extensive textual prompts, such as Princess Diana and Prince Charles for human description in Fig.~\ref{fig:wild_demo}-(a) and
Döner Kebap for object description in Fig.~\ref{fig:wild_demo}-(d). Also, as in Fig.~\ref{fig:wild_visual_demo}, \emph{MP-HOI} could detect a specific HOI composition via visual prompts, and has the flexibility to integrate the visual prompt for
interaction definition and textual prompt for object definitions (e.g., woman and phone) for the test as in Fig.~\ref{fig:wild_visual_demo}-(e).
\vspace{-0.45em}

\section{Conclusion}

We introduce \emph{MP-HOI}, a powerful multi-modal prompt-based HOI detector. It leverages both textual descriptions and visual exemplars to realize HOI detection in the open world. We introduce a large-scale unified HOI dataset named \emph{Magic-HOI}, and a high-quality synthetic HOI dataset called \emph{SynHOI} for effective training. Our results demonstrate that the trained \emph{MP-HOI}, as a generalist HOI detector, exhibits remarkable zero-shot capability in real-world scenarios and consistently achieves a new state-of-the-art performance across various benchmarks. We hope our models and datasets can serve the community for further research.

\section*{Acknowledgment}
The work is partially supported by the Young Scientists Fund of the National Natural Science Foundation of China under grant No.62106154, by the Natural Science Foundation of Guangdong Province, China (General Program) under grant No.2022A1515011524, and by Shenzhen Science and Technology Program JCYJ20220818103001002 and ZDSYS20211021111415025, and by the Guangdong Provincial Key Laboratory of Big Data Computing, The Chinese University of Hong Kong (Shenzhen).
{
    \small
    \bibliographystyle{ieeenat_fullname}
    \bibliography{main}
}


\end{document}